\documentclass{article}

\usepackage{microtype}
\usepackage{graphicx}
\usepackage{booktabs} 
\usepackage{caption}
\usepackage{subcaption}
\usepackage{makecell}
\usepackage{multirow}

\usepackage{amsmath}
\usepackage{amssymb}
\usepackage{amsfonts}
\usepackage{amsthm}

\usepackage{color}
\usepackage{xcolor,colortbl}
\definecolor{matlab-blue}{rgb}{0,0.4470,0.7410}
\definecolor{matlab-orange}{rgb}{0.8500,0.3250,0.0980}
\definecolor{matlab-yellow}{rgb}{0.9290,0.6940,0.1250}
\definecolor{matlab-green}{rgb}{0.4660,0.6740,0.1880}
\definecolor{matlab-red}{rgb}{0.6350,0.0780,0.1840}
\definecolor{matlab-purple}{rgb}{0.4901,0.1803,0.5529}
\definecolor{matlab-light-blue}{rgb}{0.298,0.741,0.929}
\definecolor{ourmethod}{gray}{0.93}
\DeclareMathOperator*{\argmax}{arg\,max}

\newcommand{\btheta}{\boldsymbol{\theta}}
\newcommand{\by}{\mathbf{y}}
\newcommand{\bx}{\mathbf{x}}

\newtheoremstyle{thm}
  {2.0pt} 
  {2.0pt} 
  {\itshape} 
  {} 
  {\bfseries} 
  {.} 
  {.5em} 
  {} 
\theoremstyle{thm}
\newtheorem{assumption}{Assumption}

\usepackage{tikz}
\usetikzlibrary{matrix}
\usetikzlibrary{calc}
\usetikzlibrary{positioning}
\usetikzlibrary{graphs}
\usetikzlibrary{shapes.geometric}
\usetikzlibrary{backgrounds}

\usepackage{enumitem}

\usepackage{alphalph}
\usepackage{pgffor}

\usepackage[decisionutilitycolor, chamferedutility]{include/influence-diagrams}

\usepackage[colorlinks]{hyperref}

\usepackage[accepted]{include/icml2020}

\icmltitlerunning{Can Autonomous Vehicles Identify, Recover From, and Adapt to Distribution Shifts?}

\usepackage[linesnumbered,ruled,noend]{algorithm2e}

\newcommand{\bftab}{\fontseries{b}\selectfont}

\begin{document}

\twocolumn[
\icmltitle{Can Autonomous Vehicles Identify, Recover From,\\and Adapt to Distribution Shifts?}



\icmlsetsymbol{equal}{*}

\begin{icmlauthorlist}
\icmlauthor{Angelos Filos}{equal,ox}
\icmlauthor{Panagiotis Tigas}{equal,ox}
\icmlauthor{Rowan McAllister}{ucb}
\icmlauthor{Nicholas Rhinehart}{ucb}
\icmlauthor{Sergey Levine}{ucb}
\icmlauthor{Yarin Gal}{ox}
\end{icmlauthorlist}

\icmlaffiliation{ox}{University of Oxford}
\icmlaffiliation{ucb}{University of California, Berkeley}

\icmlcorrespondingauthor{Angelos Filos}{angelos.filos@cs.ox.ac.uk}
\icmlcorrespondingauthor{Panagiotis Tigas}{ptigas@robots.ox.ac.uk}

\icmlkeywords{
  Autonomous Driving,
  Machine Learning,
  Reinforcement Learning,
  Online Learning,
  Self-Supervised Learning,
  Imitation Learning,
  Multi-Agent Systems,
  Benchmark,
}

\vskip 0.3in
]


\printAffiliationsAndNotice{\icmlEqualContribution} 

\begin{abstract}
Out-of-training-distribution (OOD) scenarios are a common challenge of learning agents at deployment, typically leading to arbitrary deductions and poorly-informed decisions. In principle, detection of and adaptation to OOD scenes can mitigate their adverse effects.
In this paper, we highlight the limitations of current approaches to novel driving scenes and propose an epistemic uncertainty-aware planning method, called \emph{robust imitative planning} (RIP).
Our method can detect and recover from some distribution shifts, reducing the overconfident and catastrophic extrapolations in OOD scenes.
If the model's uncertainty is too great to suggest a safe course of action, the model can instead query the expert driver for feedback, enabling sample-efficient online adaptation, a variant of our method we term \emph{adaptive robust imitative planning} (AdaRIP).
Our methods outperform current state-of-the-art approaches in the nuScenes \emph{prediction} challenge, but since no benchmark evaluating OOD detection and adaption currently exists to assess \emph{control}, we introduce an autonomous car novel-scene benchmark, \texttt{CARNOVEL}, to evaluate the robustness of driving agents to a suite of tasks with distribution shifts, where our methods outperform all the baselines.
\vspace{-2.0em}
\end{abstract}

\begin{figure}[ht]
  \centering
  \begin{subfigure}[b]{0.4\linewidth}
    \centering
    \includegraphics[width=\linewidth]{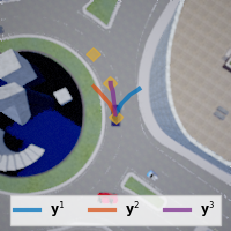}
  \caption{OOD driving scenario}
  \label{fig:didactic-scenario}
  \end{subfigure}
  \begin{subfigure}[b]{0.58\linewidth}
    \centering
    \resizebox{\linewidth}{!}{\begin{tikzpicture}
      \tikzstyle{empty}=[]
      \draw[gray!75, fill=gray!5, very thick] (0,0) rectangle (3,3);
      \draw[fill=matlab-yellow!10, dashed] (0,-1.25) rectangle (3,-0.25);
      \draw[gray!100, thick] (0,1) -- (3, 1);
      \draw[gray!100, thick] (0,2) -- (3, 2);
      \draw[gray!100, thick] (1,0) -- (1, 3);
      \draw[gray!100, thick] (2,0) -- (2, 3);
      \node[rotate=90, fill=gray!10] (E) at (-1.15, 1.50) {models, $q_{k}$};
      \node[fill=gray!10] (T) at (1.50, 4.00) {trajectories, $\by_{i}$};
      \node (T1) at (0.50, 3.35) {\color{matlab-blue}$\by^{1}$};
      \node (T2) at (1.50, 3.35) {\color{matlab-orange}$\by^{2}$};
      \node (T3) at (2.50, 3.35) {\color{matlab-purple}$\by^{3}$};
      \node (E1) at (-0.4, 2.50) {$q_{1}$};
      \node (p11) at (0.50, 2.50) {\textbf{0.6}};
      \node (p12) at (1.50, 2.50) {0.1};
      \node (p13) at (2.50, 2.50) {0.3};
      \node (E2) at (-0.4, 1.50) {$q_{2}$};
      \node (p21) at (0.50, 1.50) {0.3};
      \node (p22) at (1.50, 1.50) {\textbf{0.4}};
      \node (p23) at (2.50, 1.50) {0.3};
      \node (E3) at (-0.4, 0.50) {$q_{3}$};
      \node (p31) at (0.50, 0.50) {0.2};
      \node (p32) at (1.50, 0.50) {0.2};
      \node (p33) at (2.50, 0.50) {\textbf{0.6}};
      \node (min_label) at (-0.75, -0.75) {$\min_{k}$};
      \node (min_T1)    at (0.5, -0.75) {0.2};
      \node (min_T2)    at (1.5, -0.75) {0.1};
      \node (min_T3)    at (2.5, -0.75) {\textbf{0.3}};
      \node (maxmin_s)     at (3.0, -0.75) {};
      \node (maxmin_e)     at (4.5, -0.75) {};
      \path (maxmin_s) edge [->, black] (maxmin_e);
      \node[fill=matlab-green!10] (maxmin) at (5, -0.75) {\color{matlab-purple}$\by^{3}$};
      \node (max___) at (3.75, 3.0) {$\argmax_{i}$};
      \node (max_s1)     at (3.0, 2.5) {};
      \node (max_e1)     at (4.5, 2.5) {};
      \path (max_s1) edge [->, black] (max_e1);
      \node (max_s)     at (3.0, 1.5) {};
      \node (max_e)     at (4.5, 1.5) {};
      \path (max_s) edge [->, black] (max_e);
      \node (max_s3)     at (3.0, 0.5) {};
      \node (max_e3)     at (4.5, 0.5) {};
      \path (max_s3) edge [->, black] (max_e3);
      \node[fill=matlab-green!10] (max_M3)    at (5, 0.5) {\color{matlab-purple}$\by^{3}$};
      \node[fill=matlab-red!10] (max_M2)    at (5, 1.5) {\color{matlab-orange}$\by^{2}$};
      \node[fill=matlab-red!10] (max_M1)  at (5, 2.5) {\color{matlab-blue}$\by^{1}$};
    \end{tikzpicture}}
  \caption{Robust imitative planning}
  \label{fig:expert-likelihoods}
  \end{subfigure}
  \caption{
    Didactic example: (a) in a novel, out-of-training distribution (OOD) driving scenario, candidate plans/trajectories ${\color{matlab-blue}\by^{1}}, {\color{matlab-orange}\by^{2}}, {\color{matlab-purple}\by^{3}}$ are (b) evaluated (row-wise) by an ensemble of expert-likelihood models $q_{1}, q_{2}, q_{3}$.
    Under models $q_{1}$ and $q_{2}$ the best plans are the catastrophic trajectories ${\color{matlab-blue}\by^{1}}$ and ${\color{matlab-orange}\by^{2}}$ respectively.
    Our epistemic uncertainty-aware robust (RIP) planning method aggregates the evaluations of the ensemble and proposes the safe plan ${\color{matlab-purple}\by^{3}}$.
    RIP considers the disagreement between the models and avoid overconfident but catastrophic extrapolations in OOD tasks.
  }
  \label{fig:cartoon}
\vspace{-2.0em}
\end{figure}

\begin{figure*}
  \centering
  \begin{subfigure}[b]{0.33\linewidth}
    \centering
    \resizebox{!}{4.0cm}{
    \begin{tikzpicture}
      \tikzstyle{empty}=[]
      \node (xs_1) at (0.0, 2.25) {\includegraphics[width=.35\textwidth]{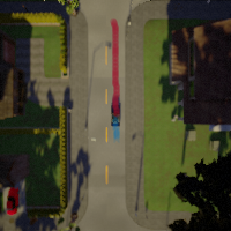}};
      \node (xs_i) at (0.0, 0.0) {\includegraphics[width=.35\textwidth]{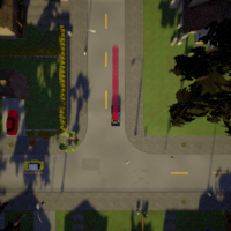}};
      \node (xs_ip1) at (2.25, 0.0) {\includegraphics[width=.35\textwidth]{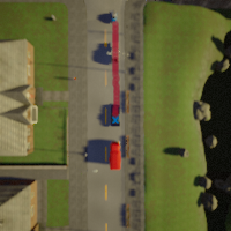}};
      \node (xs_N) at (2.25, 2.25) {\includegraphics[width=.35\textwidth]{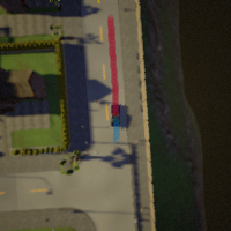}};
    \end{tikzpicture}
    }
    \caption{Demonstrations}
    \label{fig:demonstrations}
  \end{subfigure}
  \begin{subfigure}[b]{0.33\linewidth}
    \centering
    \resizebox{\linewidth}{!}{
    \begin{tikzpicture}
      \tikzstyle{empty}=[]
      \node (bootstrap) at (1.9, 4.25) {$\mathcal{D} = \mathcal{D}_  {1} \cup \cdots \mathcal{D}_{k} \cdots \cup \mathcal{D}_{K}$};
      \node (mle) at (1.9, 3.75) {$\btheta_{k, \text{MLE}} = \argmax_{\btheta} \mathbb{E}_{(\bx, \by) \sim \mathcal{D}_{k}}\left[ \log q(\by \vert \bx ; \btheta) \right]$};
      \node (x_1) at (0.0, 2.75) {$\bx \sim \mathcal{D}_{1}$};
      \draw [gray,fill=gray!10] plot coordinates {(1.5, 3.25) (2.0, 3.0) (2.0, 2.5) (1.5, 2.25)} -- cycle;
      \node[circle, minimum size=2pt] (e_1_theta) at (1.75, 2.75) {$\btheta_{1}$};
      \path (x_1) edge [->, black] (e_1_theta);
      \node (logq_1) at (3.75, 2.75) {$q(\by \vert \bx; \btheta_{1})$};
      \path (e_1_theta) edge [->, black] (logq_1);
      \node (s_1) at (0.75, 2.0) {$\by \sim \mathcal{D}_{1}$};
      \draw[->, to path={-| (\tikztotarget)}] (s_1) edge (e_1_theta);
      \node (x_K) at (0.0, 1.0) {$\bx \sim \mathcal{D}_{K}$};
      \draw [gray,fill=gray!10] plot coordinates {(1.5, 1.5) (2.0, 1.25) (2.0, 0.75) (1.5, 0.5)} -- cycle;
      \node[circle, minimum size=2pt] (e_K_theta) at (1.75, 1.0) {$\btheta_{K}$};
      \path (x_K) edge [->, black] (e_K_theta);
      \node (logq_K) at (3.75, 1.0) {$q(\by \vert \bx; \btheta_{K})$};
      \path (e_K_theta) edge [->, black] (logq_K);
      \node (s_K) at (0.75, 0.25) {$\by \sim \mathcal{D}_{K}$};
      \draw[->, to path={-| (\tikztotarget)}] (s_K) edge (e_K_theta);
      \draw[dotted] (logq_1) -- (logq_K);
      \draw[dotted] (x_1) -- (x_K);
    \end{tikzpicture}
    }
    \caption{Ensemble training}
    \label{fig:ensemble-imitative-models}
  \end{subfigure}
  \begin{subfigure}[b]{0.33\linewidth}
    \centering
    \resizebox{!}{4.0cm}{
    \begin{tikzpicture}
      \tikzstyle{empty}=[]
      \node (x_label) at (2.0, 5.0) {$\bx$};
      \node (x) at (2.0, 4.3) {\includegraphics[width=1cm]{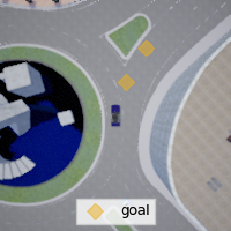}};
      \draw [gray,fill=matlab-yellow!20] plot coordinates {(1.25, 2.25) (0.75, 2.25) (0.5, 2.75) (1.5, 2.75)} -- cycle;
      \node[circle, minimum size=2pt] (e_1_theta) at (1.0, 2.5) {$\btheta_{1}$};
      \draw[->, to path={(\tikztostart.west) -| (\tikztotarget.north)}] (x) edge (e_1_theta);
      \node (s_1) at (0.25, 3.5) {\color{matlab-red}$\by$};
      \draw[->, to path={(\tikztostart.south) |- (\tikztotarget.west)}] (s_1) edge (e_1_theta);
      \draw [gray,fill=matlab-yellow!20] plot coordinates {(3.25, 2.25) (2.75, 2.25) (2.5, 2.75) (3.5, 2.75)} -- cycle;
      \node[circle, minimum size=2pt] (e_K_theta) at (3.0, 2.5) {$\btheta_{K}$};
      \draw[->, to path={(\tikztostart.east) -| (\tikztotarget.north)}] (x) edge (e_K_theta);
      \node (s_K) at (2.25, 3.5) {\color{matlab-red}$\by$};
      \draw[->, to path={(\tikztostart.south) |- (\tikztotarget.west)}] (s_K) edge (e_K_theta);
      \node [draw=gray, minimum size=2pt,fill=matlab-green!20] (aggregator) at (2.0, 1.25) {$\oplus$ e.g., $\min_{k}$, $\frac{1}{K} \sum_{k}$};
      \draw[->, to path={(\tikztostart.south) -- ([xshift=-1.0cm]\tikztotarget.north)}] (e_1_theta) edge (aggregator);
      \draw[->, to path={(\tikztostart.south) -- ([xshift=+1.0cm]\tikztotarget.north)}] (e_K_theta) edge (aggregator);
      \node (objective) at (2.0, 0.0) {$\mathcal{U} = \oplus \log q({\color{matlab-red}\by} \vert \bx, \mathcal{G}; \btheta),\ {\color{matlab-red}\frac{\partial \mathcal{U}}{\partial \by}}$};
      \node (planning) at (2.0, -0.65) {\textbf{plan} $\by^{*}= \argmax_{{\color{matlab-red}\by}} \mathcal{U}$};
      \draw[->] (aggregator) edge (objective);
      \draw[->, matlab-red, dashed, to path={([yshift=+0.2cm]\tikztostart.west) -| ([xshift=+0.2cm]\tikztotarget.south)}] (e_1_theta) edge (s_1);
      \draw[->, matlab-red, dashed, to path={([yshift=+0.2cm]\tikztostart.west) -| ([xshift=+0.2cm]\tikztotarget.south)}] (e_K_theta) edge (s_K);
      \draw[->, matlab-red, dashed, to path={([xshift=+0.2cm]\tikztostart.north) -- ([xshift=+0.2cm]\tikztotarget.south)}] (objective) edge (aggregator);
      \draw[->, matlab-red, dashed, to path={([xshift=-0.8cm]\tikztostart.north) -- ([xshift=+0.2cm]\tikztotarget.south)}] (aggregator) edge (e_1_theta);
      \draw[->, matlab-red, dashed, to path={([xshift=+1.2cm]\tikztostart.north) -- ([xshift=+0.2cm]\tikztotarget.south)}] (aggregator) edge (e_K_theta);
    \end{tikzpicture}
    }
    \caption{Planning under uncertainty}
    \label{fig:planning}
  \end{subfigure}
  \caption{
    The robust imitative planning (RIP) framework. \textbf{(a) Expert demonstrations}. We assume access to observations $\bx$ and expert state $\by$ pairs, collected either in simulation~\citep{dosovitskiy2017carla} or in real-world~\citep{caesar2019nuscenes, sun2019scalability, lyft2019}. \textbf{(b) Learning algorithm (cf. Section~\ref{subsec:bayesian-imitative-model})}. We capture epistemic model uncertainty by training an ensemble of density estimators $\{q(\by \vert \bx ; \btheta_{k})\}_{k=1}^{K}$, via maximum likelihood. Other approximate Bayesian deep learning methods~\citep{gal2016dropout} are also tested. \textbf{(c) Planning paradigm (cf. Section~\ref{subsec:planning-under-epistemic-uncerainty})}. The epistemic uncertainty is taken into account at planning via the aggregation operator $\oplus$ (e.g., $\min_{k}$), and the optimal plan $\by^{*}$ is calculated online with gradient-based optimization \emph{through the learned likelihood models}.
  }
  \label{fig:algorithm-schematic}
\vspace{-1.25em}
\end{figure*}

\section{Introduction}
\label{sec:introduction}

Autonomous agents hold the promise of systematizing decision-making to reduce catastrophes due to human mistakes.
Recent advances in machine learning (ML) enable the deployment of such agents in challenging, real-world, safety-critical domains, such as autonomous driving (AD) in urban areas.
However, it has been repeatedly demonstrated that the reliability of ML models degrades radically when they are exposed to novel settings (i.e., \emph{under a shift away from the distribution of observations seen during their training}) due to their failure to generalise, leading to catastrophic outcomes~\citep{sugiyama2012machine, amodei2016concrete, snoek2019can}.
The diminishing performance of ML models to out-of-training distribution (OOD) regimes is concerning in life-critical applications, such as AD~\citep{quionero2009dataset, leike2017ai}.

Although there are relatively simple strategies (e.g., stay within the lane boundaries, avoid other cars and pedestrians) that generalise, perception-based, end-to-end approaches, while flexible, they are also susceptible to spurious correlations. Therefore, they can pick up non-causal features that lead to confusion in OOD scenes~\citep{de2019causal}.

Due to the complexity of the real-world and its ever-changing dynamics, the deployed agents inevitably face novel situations and should be able to cope with them, to at least (a) identify and ideally (b) recover from them, without failing catastrophically.
These desiderata are \emph{not} captured by the existing benchmarks~\citep{carlachallenge, codevilla2019exploring} and as a consequence, are \emph{not} satisfied by the current state-of-the-art methods~\citep{chen2019learning, tang2019worst,rhinehart2020deep}, which are prone to fail in unpredictable ways when they experience OOD scenarios (depicted in Figure~\ref{fig:cartoon} and empirically verified in Section~\ref{sec:benchmarking-robustness-to-novelty}).

In this paper, we demonstrate the practical importance of OOD detection in AD and its importance for safety.
The key contributions are summarised as follows:
\begin{enumerate}[noitemsep]
\vspace{-1em}
  \item \textbf{Epistemic uncertainty-aware planning:}
  We present an epistemic uncertainty-aware planning method, called \emph{robust imitative planning} (RIP) for detecting and recovering from distribution shifts.
  Simple quantification of epistemic uncertainty with deep ensembles enables detection of distribution shifts.
  By employing Bayesian decision theory and robust control objectives, we show how we can act conservatively in unfamiliar states which often allows us to recover from distribution shifts (didactic example depicted in Figure~\ref{fig:cartoon}).
  \item \textbf{Uncertainty-driven online adaptation:}
  Our adaptive, online method, called \emph{adaptive robust imitative planning} (AdaRIP), uses RIP's epistemic uncertainty estimates to efficiently query the expert for feedback which is used to adapt on-the-fly, without compromising safety.
  Therefore, AdaRIP could be deployed in the real world: it can reason about what it does not know and in these cases ask for human guidance to guarantee current safety and enhance future performance.
  \item \textbf{Autonomous car novel-scene benchmark:}
  We introduce an autonomous car novel-scene benchmark, called \texttt{CARNOVEL}, to assess the robustness of AD methods to a suite of out-of-distribution tasks.
  In particular, we evaluate them in terms of their ability to: (a) detect OOD events, measured by the correlation of infractions and model uncertainty; (b) recover from distribution shifts, quantified by the percentage of successful manoeuvres in novel scenes and (c) efficiently adapt to OOD scenarios, provided online supervision.
\end{enumerate}

\section{Problem Setting and Notation}
\label{sec:problem-setting-and-notation}

We consider sequential decision-making in safety-critical domains.
A method is considered safety when it is accurate, with respect to some metric (cf. Sections~\ref{sec:benchmarking-robustness-to-novelty},~\ref{sec:benchmarking-adaptation}), and certain.
\begin{assumption}[Expert demonstrations]
  We assume access to a dataset, $\mathcal{D}=\{(\bx^{i}, \by^{i})\}_{i=1}^{N}$, of time-profiled expert trajectories (i.e., plans), $\by$, paired with high-dimensional observations, $\bx$, of the corresponding scenes.
  The trajectories are drawn from the expert policy, $\by \sim \pi_{\text{expert}}(\cdot \vert \bx)$.
\label{ass:expert-demonstrations}
\end{assumption}
Our goal is to approximate the (i.e., near-optimal) unknown expert policy, $\pi_{\text{expert}}$, using imitation learning~\citep[IL]{widrow1964pattern, pomerleau1989alvinn}, based only on the demonstrations, $\mathcal{D}$.
For simplicity, we also make the following assumptions, common in the autonomous driving and robotics literature~\citep{rhinehart2020deep, du2019model}.
\begin{assumption}[Inverse dynamics]
  We assume access to an inverse dynamics model~\citep[PID controller, $\mathbb{I}$]{bellman2015adaptive}, which performs the low-level control -- inverse planning -- $a_{t}$ (i.e., steering, braking and throttling), provided the current and next states (i.e., positions), $s_{t}$ and $s_{t+1}$, respectively. Therefore, we can operate directly on state-only trajectories, $\by=\left(s_{1}, \ldots, s_{T}\right)$, where the actions are determined by the local planner, $a_{t} = \mathbb{I}(s_{t}, s_{t+1}),\ \forall t=1,\ldots,T-1$.
\label{ass:local-controller}
\end{assumption}
\begin{assumption}[Global planner]
  We assume access to a global navigation system that we can use to specify high-level goal locations $\mathcal{G}$ or/and commands $\mathcal{C}$ (e.g., turn left/right at the intersection, take the second exit).
\label{ass:global-planner}
\end{assumption}
\begin{assumption}[Perfect localization]
  We consider the provided locations (e.g., goal, ego-vehicle positions) as accurate, i.e., filtered by a localization system.
\label{ass:perfect-localization}
\end{assumption}
These are benign assumptions for many applications in robotics.
If required, these quantities can also be learned from data, and are typically easier to learn than $\pi_{\text{expert}}$.

\section{Robust Imitative Planning}
\label{sec:robust-imitative-planning}
We seek an imitation learning method that (a) provides a distribution over expert plans; (b) quantifies epistemic uncertainty to allow for detection of OOD observations and (c) enables robustness to distribution shift with an explicit mechanism for recovery.
Our method is shown in Figure~\ref{fig:algorithm-schematic}.
First, we present the model used for imitating the expert.

\subsection{Bayesian Imitative Model}
\label{subsec:bayesian-imitative-model}
We perform context-conditioned density estimation of the distribution over future expert trajectories (i.e., plans), using a probabilistic ``imitative'' model $q(\by \vert \bx; \btheta)$, trained via maximum likelihood estimation (MLE):
%
\begin{align}
  \btheta_{\text{MLE}} = \argmax_{\btheta} \mathbb{E}_{(\bx, \by) \sim \mathcal{D}}\left[ \log q(\by \vert \bx ; \btheta) \right].
\label{eq:mle}
\end{align}
Contrary to existing methods in AD~\citep{rhinehart2020deep, chen2019learning}, we place a prior distribution $p(\btheta)$ over possible model parameters $\btheta$, which induces a distribution over the density models $q(\by \vert \bx; \btheta)$.
After observing data $\mathcal{D}$, the distribution over density models has a posterior $p(\btheta \vert \mathcal{D})$.

\textbf{Practical implementation.}
~We use an autoregressive neural density estimator~\cite{rhinehart2018r2p2}, depicted in Figure~\ref{fig:ensemble-imitative-models}, as the imitative model, parametrised by learnable parameters $\btheta$. The likelihood of a plan $\by$ in context $\bx$ to come from an expert (i.e., \emph{imitation prior}) is given by:
%
\begin{align}
  q(\by \vert \bx ; \btheta)
    &= \prod_{t=1}^{T} p(s_{t} \vert \by_{<t}, \bx; \btheta)\nonumber\\
    &= \prod_{t=1}^{T} \mathcal{N}(s_{t}; \mu(\by_{<t}, \bx; \btheta), \Sigma(\by_{<t}, \bx; \btheta)),
\label{eq:r2p2}
\end{align}
where $\mu(\cdot; \btheta)$ and $\Sigma(\cdot; \btheta)$ are two heads of a recurrent neural network, with shared torso.
We decompose the imitation prior as a telescopic product (cf. Eqn.~(\ref{eq:r2p2})), where conditional densities are assumed normally distributed, and the distribution parameters are learned (cf. Eqn.~(\ref{eq:mle})).
Despite the unimodality of normal distributions, the autoregression (i.e., sequential sampling of normal distributions where the future samples depend on the past) allows to model multi-modal distributions~\citep{uria2016neural}.
Although more expressive alternatives exist, such as the mixture of density networks~\citep{bishop1994mixture} and normalising flows~\citep{rezende2015variational}, we empirically find Eqn.~(\ref{eq:r2p2}) sufficient.

The estimation of the posterior of the model parameters, $p(\btheta \vert \mathcal{D})$, with exact inference is intractable for non-trivial models~\citep{neal2012bayesian}.
We use ensembles of deep imitative models as a simple approximation to the posterior $p(\btheta \vert \mathcal{D})$.
We consider an ensemble of $K$ components, using $\btheta_{k}$ to refer to the parameters of our $k$-th model $q_{k}$, trained with via maximum likelihood (cf. Eqn.~(\ref{eq:mle}) and Figure~\ref{fig:ensemble-imitative-models}).
However, any (approximate) inference method to recover the posterior $p(\btheta \vert \mathcal{D})$ would be applicable.
To that end, we also try Monte Carlo dropout~\citep{gal2016dropout}.

\subsection{Detecting Distribution Shifts}
The log-likelihood of a plan $\log q(\by \vert \bx;\btheta)$ (i.e., \emph{imitation prior}) is a proxy of the quality of a plan $\by$ in context $\bx$ under model $\btheta$.
We detect distribution shifts by looking at the disagreement of the qualities of a plan under models coming from the posterior, $p(\btheta \vert \mathcal{D})$.
We use the variance of the imitation prior with respect to the model posterior, i.e.,
%
\begin{align}
  u(\by) \triangleq \text{Var}_{p(\btheta \vert \mathcal{D})}\left[ \log q(\by \vert \bx; \btheta) \right]
\label{eq:var}
\end{align}
to quantify the model disagreement: Plans at in-distribution scenes have low variance, but high variance in OOD scenes.
We can efficiently calculate Eqn.~(\ref{eq:var}) when we use ensembles, or Monte Carlo, sampling-based methods for $p(\btheta \vert \mathcal{D})$.

Having to commit to a decision, just the detection of distribution shifts via the quantification of epistemic uncertainty is insufficient for recovery.
Next, we introduce an epistemic uncertainty-aware planning objective that allows for robustness to distribution shifts.


\subsection{Planning Under Epistemic Uncertainty}
\label{subsec:planning-under-epistemic-uncerainty}
We formulate planning to a goal location $\mathcal{G}$ under epistemic uncertainty, i.e.,\ posterior over model parameters $p(\btheta \vert \mathcal{D})$, as the optimization~\citep{barber2012bayesian} of the generic objective, which we term \emph{robust imitative planning} (RIP):

{\small
\begin{align}
  \by^{\mathcal{G}}_{\text{RIP}}
    \!&\triangleq\! \argmax_{\by} \overbrace{\underset{\btheta \in \text{supp}\big(p(\btheta \vert \mathcal{D})\big)}{\oplus}}^{\text{aggregation operator}} \; \log \underbrace{p(\by \vert \mathcal{G}, \bx;\btheta)}_{\text{imitation posterior}} \nonumber \\
    \!&=\!\argmax_{\by} \underset{\btheta \in \text{supp}\big(p(\btheta \vert \mathcal{D})\big)}{\oplus} \! \log \!\underbrace{q(\by \vert \bx;\btheta)}_{\text{imitation prior}} + \log \!\!\!\!\! \underbrace{p(\mathcal{G} \vert \by)}_{\text{goal likelihood\quad}},
\label{eq:generic-objective}
\end{align}
}%
where $\oplus$ is an operator (defined below) applied on the posterior $p(\btheta \vert \mathcal{D})$ and the goal-likelihood is given, for example, by a Gaussian centred at the final goal location $s_{T}^{\mathcal{G}}$ and a pre-specified tolerance $\epsilon$, $p(\mathcal{G} \vert \by) = \mathcal{N}(\by_{T}; \by_{T}^{\mathcal{G}}, \epsilon^{2} I)$.

Intuitively, we choose the plan $\by^{\mathcal{G}}_{\text{RIP}}$ that maximises the likelihood to have come from an expert demonstrator (i.e., ``imitation prior'') and is ``close'' to the goal $\mathcal{G}$.
The model posterior $p(\btheta \vert \mathcal{D})$ represents our belief (uncertainty) about the true expert model, having observed data $\mathcal{D}$ and from prior $p(\btheta)$ and the aggregation operator $\oplus$ determines our level of awareness to uncertainty under a unified framework.

For example, a deep imitative model~\citep{rhinehart2020deep} is a particular instance of the more general family of objectives described by Eqn.~(\ref{eq:generic-objective}), where the operator $\oplus$ selects a single $\btheta_{k}$ from the posterior (point estimate).
However, this approach is oblivious to the epistemic uncertainty and prone to fail in unfamiliar scenes (cf. Section~\ref{sec:benchmarking-robustness-to-novelty}).

In contrast, we focus our attention on two aggregation operators due to their favourable properties, which take epistemic uncertainty into account: (a) one inspired by robust control~\citep{wald1939contributions} which encourages pessimism in the face of uncertainty and (b) one from Bayesian decision theory, which marginalises the epistemic uncertainty.
Table~\ref{tab:methods} summarises the different operators considered in our experiments. Next, we motivate the used operators.

\subsubsection{Worst Case Model (RIP-WCM)}
\label{subsubsec:worst-case-model}

In the face of (epistemic) uncertainty, robust control~\citep{wald1939contributions} suggests to act pessimistically -- reason about the \emph{worst case scenario} and optimise it.
All models with non-zero posterior probability $p(\btheta \vert \mathcal{D})$ are likely and hence our \emph{robust imitative planning with respect to the worst case model} (RIP-WCM) objective acts with respect to the most pessimistic model, i.e.,
%
\begin{align}
  s_{\text{RIP-WCM}}
  \;&\triangleq\; \argmax_{\by} {\color{matlab-red}\min_{\btheta \in \text{supp}\big(p(\btheta \vert \mathcal{D})\big)}} \log q(\by \vert \bx;\btheta) \,.
\label{eq:RIP-WCM}
\end{align}
%
The solution of the $\argmax_{\by} \min_{\btheta}$ optimization problem in Eqn.~(\ref{eq:RIP-WCM}) is generally not tractable, but our deep ensemble approximation enables us to solve it by evaluating the minimum over a finite number of $K$ models. The maximization over plans, $\by$, is solved with online gradient-based adaptive optimization, specifically ADAM~\citep{kingma2014adam}.
An alternative online planning method with a trajectory library~\citep{liu2009standing} (c.f. Appendix~\ref{app:onlin-planning-with-a-trajectory-library}) is used too but its performance in OOD scenes is noticeably worse than online gradient descent.

Alternative, ``softer'' robust operators can be used instead of the minimum, including the Conditional Value at Risk~\citep[CVaR]{embrechts2013modelling, rajeswaran2016epopt} that employs quantiles.
CVaR may be more useful in cases of full support model posterior, where there may be a pessimistic but trivial model, for example, due to misspecification of the prior, $p(\btheta)$, or due to the approximate inference procedure.
Mean-variance optimization~\citep{kahn2017uncertainty,kenton2019generalizing} can be also used, aiming to directly minimise the distribution shift metric, as defined in Eqn.~(\ref{eq:var}).

Next, we present a different aggregator for epistemic uncertainty that is not as pessimistic as RIP-WCM and, as found empirically, works sufficiently well too.

\subsubsection{Model Averaging (RIP-MA)}
\label{subsubsec:model-averaging}

In the face of (epistemic) uncertainty, Bayesian decision theory~\citep{barber2012bayesian} uses the predictive posterior (i.e., model averaging), which weights each model's contribution according to its posterior probability, i.e.,
%
\begin{align}
  s_{\text{RIP-MA}}
  \;&\triangleq\; \argmax_{\by} {\color{matlab-red}\int p(\btheta \vert \mathcal{D})} \log q(\by \vert \bx;\btheta) {\color{matlab-red} \mathrm{d}\btheta} \,.
\label{eq:RIP-MA}
\end{align}
%
Despite the intractability of the exact integration, the ensemble approximation used allows us to efficiently estimate and optimise the objective.
We call this method \emph{robust imitative planning with model averaging} (RIP-MA), where the more likely models' impacts are up-weighted according to the predictive posterior.

From a multi-objective optimization point of view, we can interpret the log-likelihood, $\log q(\by|\bx;\btheta)$, as the utility of a task $\btheta$, with importance $p(\btheta \vert \mathcal{D})$, given by the posterior density.
Then RIP-MA in Eqn.~(\ref{eq:RIP-MA}) gives the Pareto efficient solution~\citep{barber2012bayesian} for the tasks $\btheta \in \text{supp}\big(p(\btheta \vert \mathcal{D})\big)$.

\begin{table}[h]
  \centering
  \caption{
    Robust imitative planning (RIP) unified framework.
    The different aggregation operators applied on the posterior distribution $p(\btheta \vert \mathcal{D})$, approximated with the deep ensemble~\citep{lakshminarayanan2017simple} components $\btheta_{k}$.
  }
  \label{tab:methods}
  \resizebox{\linewidth}{!}{
  \begin{tabular}{lrr}
  \toprule
  \textbf{Methods}        & \textbf{Operator $\oplus$} & \textbf{Interpretation}     \\
  \midrule
  Imitative Models        & $\log q_{k=1}$              & Sample  \\
  Best Case (RIP-BCM)     & $\max_k \log q_k$           & Max  \\
  \midrule
  \multicolumn{3}{c}{\textbf{Robust Imitative Planning} (ours)} \\
  \midrule
  \rowcolor{ourmethod}
  Model Average (RIP-MA)  & $\sum_k \log q_k$           & Geometric Mean  \\
  \rowcolor{ourmethod}
  Worst Case (RIP-WCM)   & $\min_k \log q_k$           & Min \\
  \bottomrule
  \end{tabular}
  }
\vspace{-0.5em}
\end{table}

\begin{figure}[h]
  \centering
  \begin{subfigure}[b]{0.48\linewidth}
    \centering
    \includegraphics[width=\linewidth]{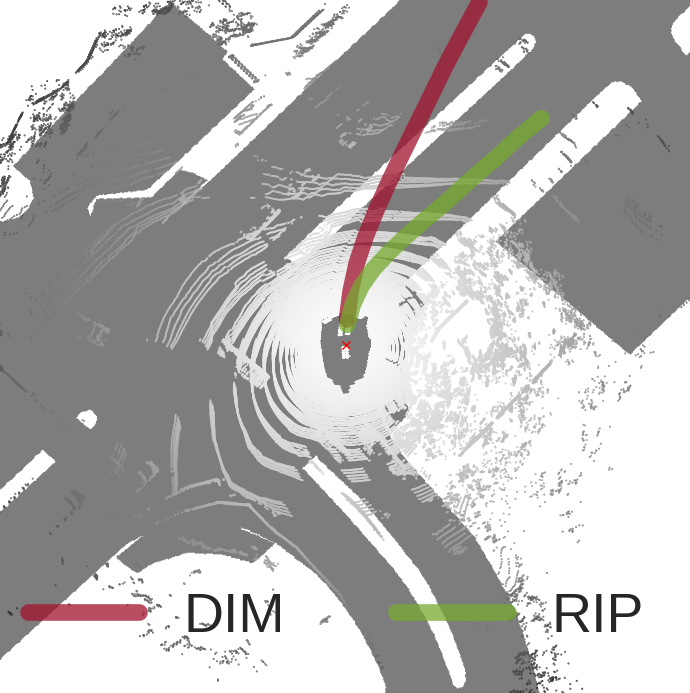}
    \caption{
      \texttt{nuScenes}
    }
  \end{subfigure}
  \begin{subfigure}[b]{0.48\linewidth}
    \centering
    \includegraphics[width=\linewidth]{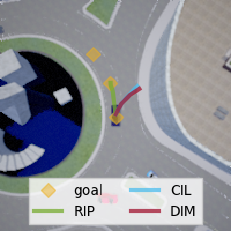}
    \caption{
      \texttt{CARNOVEL}
    }
  \end{subfigure}
\caption{
  RIP's (ours) robustness to OOD scenarios, compared to ~\citep[][CIL]{codevilla2018end} and ~\citep[][DIM]{rhinehart2020deep}.
}
\label{fig:cherries}
\vspace{-0.5em}
\end{figure}

\section{Benchmarking Robustness to Novelty}
\label{sec:benchmarking-robustness-to-novelty}

\begin{table*}[t]
  \centering
  \caption{
    We evaluate different autonomous driving prediction methods in terms of their robustness to distribution scene, in the nuScenes ICRA 2020 challenge~\citep{phan2019covernet}.
    We use the provided train--val--test splits and report performance on the test (i.e., out-of-sample) scenarios.
    A ``$\clubsuit$'' indicates methods that use LIDAR observation, as in~\citep{rhinehart2019precog}, and a ``$\diamondsuit$'' methods that use bird-view privileged information, as in~\citep{phan2019covernet}.
    A ``$\bigstar$'' indicates that we used the results from the original paper, otherwise we used our implementation.
    \textcolor{black!50}{Standard errors} are in gray (via bootstrap sampling).
    The \textbf{outperforming} method is in bold.
  }
  \label{tab:nuscenes}
  \resizebox{\linewidth}{!}{
  \begin{tabular}{llll|lll}
    \toprule
                                            &
    \multicolumn{3}{c}{\texttt{Boston}}     &
    \multicolumn{3}{c}{\texttt{Singapore}}  \\
    \cline{2-4} \cline{5-7} \\
                                    &
    minADE$_{1}$ $\downarrow$       &
    minADE$_{5}$ $\downarrow$       &
    minFDE$_{1}$ $\downarrow$       &
    minADE$_{1}$ $\downarrow$       &
    minADE$_{5}$ $\downarrow$       &
    minFDE$_{1}$ $\downarrow$       \\
    \textbf{Methods}                &
    \multicolumn{3}{c|}{($2073$ scenes, $50$ samples, open-loop planning)} &
    \multicolumn{3}{c}{($1189$ scenes, $50$ samples, open-loop planning)}  \\
    \midrule
    MTP$^{\diamondsuit \bigstar}$~\citep{cui2019multimodal} &
      $4.13$ &
      $3.24$ &
      $9.23$ &
      $4.13$ &
      $3.24$ &
      $9.23$ \\
    MultiPath$^{\diamondsuit \bigstar}$~\citep{chai2019multipath} &
      $3.89$ &
      $3.34$ &
      $9.19$ &
      $3.89$ &
      $3.34$ &
      $9.19$ \\
    CoverNet$^{\diamondsuit \bigstar}$~\citep{phan2019covernet} &
      $3.87$ &
      $2.41$ &
      $9.26$ &
      $3.87$ &
      $2.41$ &
      $9.26$ \\
    \midrule
    DIM$^{\clubsuit}$~\citep{rhinehart2020deep} &
      $3.64{\color{black!50}\pm0.05}$ &
      $2.48{\color{black!50}\pm0.02}$ &
      $8.22{\color{black!50}\pm0.13}$ &
      $3.82{\color{black!50}\pm0.04}$ &
      $2.95{\color{black!50}\pm0.01}$ &
      $8.91{\color{black!50}\pm0.08}$ \\
    RIP-BCM$^{\clubsuit}$ (baseline, cf. Table~\ref{tab:methods}) &
      $3.53{\color{black!50}\pm0.04}$ &
      $2.37{\color{black!50}\pm0.01}$ &
      $7.92{\color{black!50}\pm0.09}$ &
      $3.57{\color{black!50}\pm0.02}$ &
      $2.70{\color{black!50}\pm0.01}$ &
      $8.39{\color{black!50}\pm0.03}$ \\
    \midrule
    \rowcolor{ourmethod}
    RIP-MA$^{\clubsuit}$ (ours, cf. Section~\ref{subsubsec:model-averaging}) &
      $3.39{\color{black!50}\pm0.03}$ &
      $2.33{\color{black!50}\pm0.01}$ &
      $7.62{\color{black!50}\pm0.07}$ &
      $3.48{\color{black!50}\pm0.01}$ &
      $2.69{\color{black!50}\pm0.02}$ &
      $8.19{\color{black!50}\pm0.02}$ \\
    \rowcolor{ourmethod}
    RIP-WCM$^{\clubsuit}$ (ours, cf. Section~\ref{subsubsec:worst-case-model}) &
      \bftab{3.29}${\color{black!50}\pm0.03}$ &
      \bftab{2.28}${\color{black!50}\pm0.00}$ &
      \bftab{7.45}${\color{black!50}\pm0.05}$ &
      \bftab{3.43}${\color{black!50}\pm0.01}$ &
      \bftab{2.66}${\color{black!50}\pm0.01}$ &
      \bftab{8.09}${\color{black!50}\pm0.04}$ \\
    \bottomrule
  \end{tabular}}
\end{table*}

We designed our experiments to answer the following questions:
\textbf{Q1.} Can autonomous driving, imitation-learning, epistemic-uncertainty unaware methods detect distribution shifts?
\textbf{Q2.} How robust are these methods under distribution shifts, i.e., can they recover?
\textbf{Q3.} Does RIP's epistemic uncertainty quantification enable identification of novel scenes?
\textbf{Q4.} Does RIP's explicit mechanism for recovery from distribution shifts lead to improved performance?
To that end, we conduct experiments both on real data, in Section~\ref{sub:nuscenes}, and on simulated scenarios, in Section~\ref{sub:carnovel}, comparing our method (RIP) against current state-of-the-art driving methods.

\subsection{\texttt{nuScenes}}
\label{sub:nuscenes}
We first compare our robust planning objectives (cf. Eqn.~(\ref{eq:RIP-WCM}--\ref{eq:RIP-MA})) against existing state-of-the-art imitation learning methods in a prediction task~\citep{phan2019covernet}, based on \texttt{nuScenes}~\citep{caesar2019nuscenes}, the public, real-world, large-scale dataset for autonomous driving.
Since we do not have control over the scenes split, we cannot guarantee that the evaluation is under distribution shifts, but only test out-of-sample performance, addressing question \textbf{Q4}.

\subsubsection{Metrics}
\label{subsub:nuscenes-metrics}

For fair comparison with the baselines, we use the metrics from the ICRA 2020 \texttt{nuScenes} prediction challenge.

\textbf{Displacement error.}
~The quality of a plan, $\by$, with respect to the ground truth prediction, $\by^{*}$ is measured by the average displacement error, i.e.,
\begin{align}
  \text{ADE}(\by) \triangleq \frac{1}{T} \sum_{t=1}^{T} \Vert s_{t} - s_{t}^{*} \Vert \,,
\end{align}
where $\by=(s_{1},\ldots,s_{T})$.
Stochastic models, such as our imitative model, $q(\by \vert \bx; \btheta)$, can be evaluated based on their samples, using the minimum (over $k$ samples) ADE (i.e., minADE$_{k}$), i.e.,
\begin{align}
  \text{minADE}_{k}(q) \triangleq \min_{\{\by^{i}\}_{i=1}^{k} \sim q(\by \vert \bx)} \text{ADE}(\by^{i}) \,.
\end{align}
In prior work,~\citet{phan2019covernet} studied minADE$_{k}$ for $k>1$ in order to assess the quality of the generated samples from a model, $q$.
Although we report minADE$_{k}$ for $k=\{1, 5\}$, we are mostly interested in the decision-making (planning) task, where the driving agent commits to a \emph{single} plan, $k=1$.
We also study the final displacement error (FDE), or equivalently minFDE$_{1}$, i.e.,
\begin{align}
  \text{minFDE}_{1}(\by) \triangleq \Vert s_{T} - s_{T}^{*} \Vert \,.
\end{align}

\subsubsection{Baselines}
\label{subsub:nuscenes-baselines}

We compare our contribution to state-of-the-art methods in the \texttt{nuScenes} dataset: the Multiple-Trajectory Prediction~\citep[MTP]{cui2019multimodal}, MultiPath~\citep{chai2019multipath} and CoverNet~\citep{phan2019covernet}, all of which score a (fixed) set of trajectories, i.e., trajectory library~\citep{liu2009standing}.
Moreover, we implement the Deep Imitative Model~\citep[DIM]{rhinehart2020deep} and an optimistic variant of RIP, termed RIP-BCM and described in Table~\ref{tab:methods}.

\begin{table*}[ht]
  \centering
  \caption{
    We evaluate different autonomous driving methods in terms of their robustness to distribution shifts, in our new benchmark, \texttt{CARNOVEL}.
    All methods are trained on CARLA \texttt{Town01} using imitation learning on expert demonstrations from the autopilot~\citep{dosovitskiy2017carla}.
    A ``$\dagger$'' indicates methods that use first-person camera view, as in~\citep{chen2019learning}, a ``$\clubsuit$'' methods that use LIDAR observation, as in~\citep{rhinehart2020deep} and a ``$\diamondsuit$'' methods that use the ground truth game engine state, as in~\citep{chen2019learning}.
    A ``$\bigstar$'' indicates that we used the reference implementation from the original paper, otherwise we used our implementation.
    For all the scenes we chose pairs of start-destination locations and ran $10$ trials with randomised initial simulator state for each pair.
    \textcolor{black!50}{Standard errors} are in gray (via bootstrap sampling).
    The \textbf{outperforming} method is in bold.
    The complete \texttt{CARNOVEL} benchmark results are in Appendix~\ref{app:experimental-results-on-carnovel}.
  }
  \label{tab:carnovel}
  \resizebox{\linewidth}{!}{
  \begin{tabular}{lll|ll|ll}
    \toprule
                                                &
    \multicolumn{2}{c}{\texttt{AbnormalTurns}}  &
    \multicolumn{2}{c}{\texttt{Hills}}          &
    \multicolumn{2}{c}{\texttt{Roundabouts}}    \\
    \cline{2-3} \cline{4-5} \cline{6-7}         \\
                                    &
    Success $\uparrow$              &
    Infra/km $\downarrow$           &
    Success $\uparrow$              &
    Infra/km $\downarrow$           &
    Success $\uparrow$              &
    Infra/km $\downarrow$           \\
    \textbf{Methods}                &
    ($7 \times 10$ scenes, $\%$)    &
    ($\times 1e{-3}$)                 &
    ($4 \times 10$ scenes, $\%$)    &
    ($\times 1e{-3}$)                 &
    ($5 \times 10$ scenes, $\%$)    &
    ($\times 1e{-3}$)                 \\
    \midrule
    CIL$^{\clubsuit \bigstar}$~\citep{codevilla2018end} &
      $65.71{\color{black!50}\pm07.37}$ &
      $7.04{\color{black!50}\pm5.07}$ &
      $60.00{\color{black!50}\pm29.34}$ &
      $4.74{\color{black!50}\pm3.02}$ &
      $20.00{\color{black!50}\pm00.00}$ &
      $4.60{\color{black!50}\pm3.23}$ \\
    LbC$^{\dagger \bigstar}$~\citep{chen2019learning} &
      $00.00{\color{black!50}\pm00.00}$ &
      $5.81{\color{black!50}\pm0.58}$ &
      $50.00{\color{black!50}\pm00.00}$ &
      $1.61{\color{black!50}\pm0.15}$ &
      $08.00{\color{black!50}\pm10.95}$ &
      \bftab 3.70${{\color{black!50}\pm0.72}}$ \\
    LbC-GT$^{\diamondsuit \bigstar}$~\citep{chen2019learning} &
      $02.86{\color{black!50}\pm06.39}$ &
      \bftab 3.68${{\color{black!50}\pm0.34}}$ &
      $05.00{\color{black!50}\pm11.18}$ &
      $3.36{\color{black!50}\pm0.26}$ &
      $00.00{\color{black!50}\pm00.00}$ &
      $6.47{\color{black!50}\pm0.99}$ \\
    \midrule
    DIM$^{\clubsuit}$~\citep{rhinehart2020deep} &
      $74.28{\color{black!50}\pm11.26}$ &
      $5.56{\color{black!50}\pm4.06}$ &
      $70.00{\color{black!50}\pm10.54}$ &
      $6.87{\color{black!50}\pm4.09}$ &
      $20.00{\color{black!50}\pm09.42}$ &
      $6.19{\color{black!50}\pm4.73}$ \\
    RIP-BCM$^{\clubsuit}$ (baseline, cf. Table~\ref{tab:methods}) &
      $68.57{\color{black!50}\pm09.03}$ &
      $7.93{\color{black!50}\pm3.73}$ &
      $75.00{\color{black!50}\pm00.00}$ &
      $5.49{\color{black!50}\pm4.03}$ &
      $06.00{\color{black!50}\pm09.66}$ &
      $6.78{\color{black!50}\pm7.05}$ \\
    \midrule
    \rowcolor{ourmethod}
    RIP-MA$^{\clubsuit}$ (ours, cf. Section~\ref{subsubsec:model-averaging}) &
      \bftab 84.28${{\color{black!50}\pm14.20}}$ &
      $7.86{\color{black!50}\pm5.70}$ &
      \bftab 97.50${{\color{black!50}\pm07.90}}$ &
      \bftab 0.26${{\color{black!50}\pm0.54}}$ &
      \bftab 38.00${{\color{black!50}\pm06.32}}$ &
      $5.48{\color{black!50}\pm5.56}$ \\
    \rowcolor{ourmethod}
    RIP-WCM$^{\clubsuit}$ (ours, cf. Section~\ref{subsubsec:worst-case-model}) &
      \bftab 87.14${{\color{black!50}\pm14.20}}$ &
      $4.91{\color{black!50}\pm3.60}$ &
      \bftab 87.50${{\color{black!50}\pm13.17}}$ &
      \bftab 1.83${{\color{black!50}\pm1.73}}$ &
      \bftab 42.00${{\color{black!50}\pm06.32}}$ &
      $4.32{\color{black!50}\pm1.91}$ \\
    \bottomrule
  \end{tabular}}
\end{table*}

\subsubsection{Offline Forecasting Experiments}
\label{subsub:offline-forecasting-experiments}

We use the provided train-val-test splits from~\citep{phan2019covernet}, for towns \texttt{Boston} and \texttt{Singapore}.
For all methods we use $N=50$ trajectories, and in case of both DIM and RIP, we only optimise the ``imitation prior'' (cf. Eqn.~\ref{eq:generic-objective}), since goals are not provided, running $N$ planning procedures with different random initializations.
The performance of the baselines and our methods are reported on Table~\ref{tab:nuscenes}.
We can affirmatively answer \textbf{Q4} since RIP consistently outperforms the current state-of-the-art methods in out-of-sample evaluation.
Moreover, \textbf{Q2} can be partially answered, since the epistemic-uncertainty-unaware baselines underperformed compared to RIP.

Nonetheless, since we do not have full control over train and test splits at the ICRA 2020 challenge and hence we cannot introduce distribution shifts, we are not able to address questions \textbf{Q1} and \textbf{Q3} with the nuScenes benchmark.
To that end, we now introduce a control benchmark based on the CARLA driving simulator~\citep{dosovitskiy2017carla}.

\subsection{\texttt{CARNOVEL}}
\label{sub:carnovel}

In order to access the robustness of AD methods to novel, OOD driving scenarios, we introduce a benchmark, called \texttt{CARNOVEL}.
In particular, \texttt{CARNOVEL} is built on the CARLA simulator~\citep{dosovitskiy2017carla}.
Offline expert demonstrations\footnote{using the CARLA rule-based autopilot~\citep{dosovitskiy2017carla} without actuator noise.} from \texttt{Town01} are provided for training.
Then, the driving agents are evaluated on a suite of OOD navigation tasks, including but not limited to roundabouts, challenging non-right-angled turns and hills, none of which are experienced during training.
The \texttt{CARNOVEL} tasks are summarised in Appendix~\ref{app:carnovel}.
Next, we introduce metrics that quantify and help us answer questions \textbf{Q1}, \textbf{Q3}.

\subsubsection{Metrics}
\label{subsub:carnovel-metrics}

Since we are studying navigation tasks, agents should be able to \emph{reach safely} pre-specified \emph{destinations}.
As done also in previous work~\citep{codevilla2018end,rhinehart2020deep,chen2019learning}, the \textbf{infractions per kilometre} metric (i.e., violations of rules of the road and accidents per driven kilometre) measures how safely the agent navigates.
The \textbf{success rate} measures the percentage of successful navigations to the destination, without any infraction.
However, these standard metrics do not directly reflect the methods' performance under distribution shifts.
As a result, we introduce two new metrics for quantifying the performance in out-of-training distribution tasks:

\textbf{Detection score.}
~The correlation of infractions and model's uncertainty termed \emph{detection score} is used to measure a method's ability to predict the OOD scenes that lead to catastrophic events.
As discussed by~\citet{michelmore2018evaluating}, we look at time windows of 4 seconds~\citep{taoka1989brake, coley2009driver}.
A method that can detect potential infractions should have high detection score.

\textbf{Recovery score.}
~The percentage of successful manoeuvres in novel scenes --- where the uncertainty-unaware methods fail --- is used to quantify a method's ability to recover from distribution shifts.
We refer to this metric as \emph{recovery score}.
A method that is oblivious to novelty should have 0 recovery score, but positive otherwise.

\begin{figure*}[ht]
  \centering
  \begin{subfigure}[b]{0.23\linewidth}
    \centering
    \includegraphics[width=\textwidth]{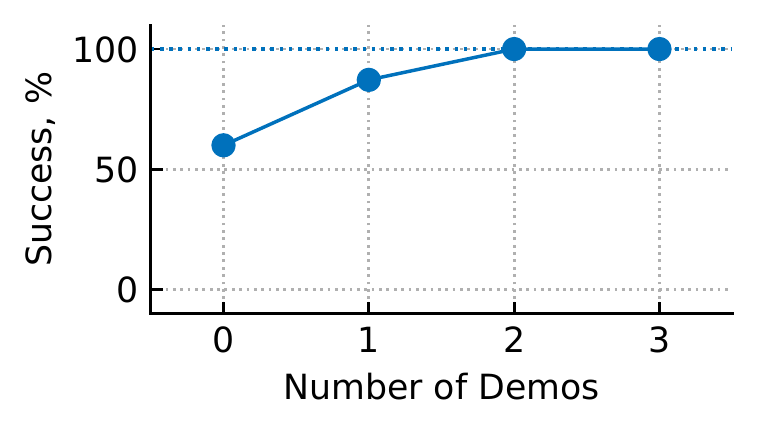}
    \caption{\texttt{AbnormalTurns4-v0}}
  \end{subfigure}
  ~
  \begin{subfigure}[b]{0.23\linewidth}
    \centering
    \includegraphics[width=\textwidth]{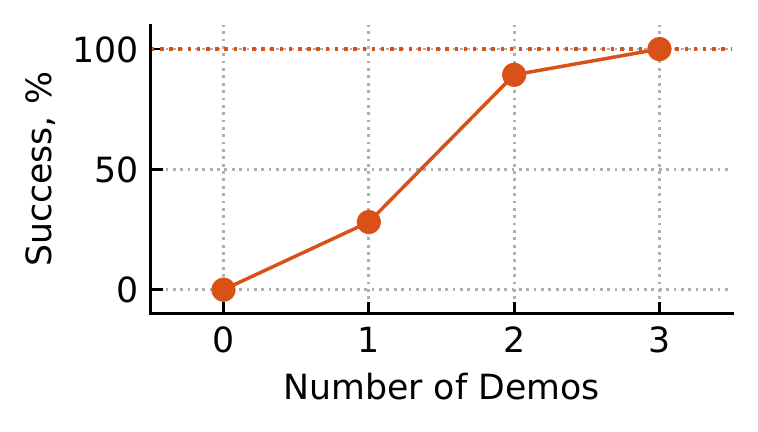}
    \caption{\texttt{BusyTown2-v0}}
  \end{subfigure}
  ~
  \begin{subfigure}[b]{0.23\linewidth}
    \centering
    \includegraphics[width=\textwidth]{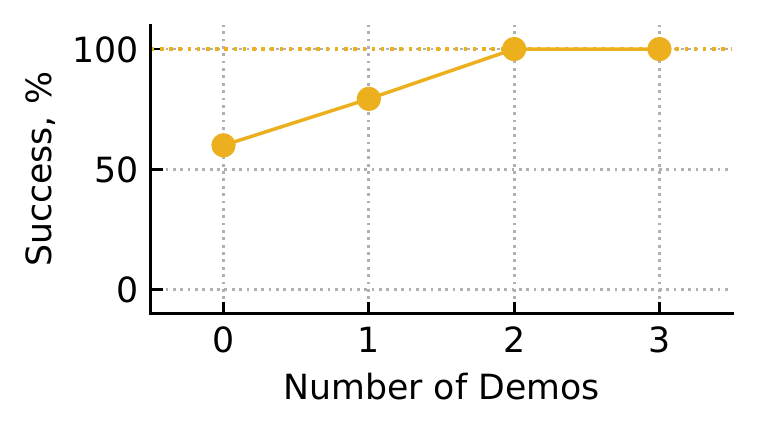}
    \caption{\texttt{Hills1-v0}}
  \end{subfigure}
  ~
  \begin{subfigure}[b]{0.23\linewidth}
    \centering
    \includegraphics[width=\textwidth]{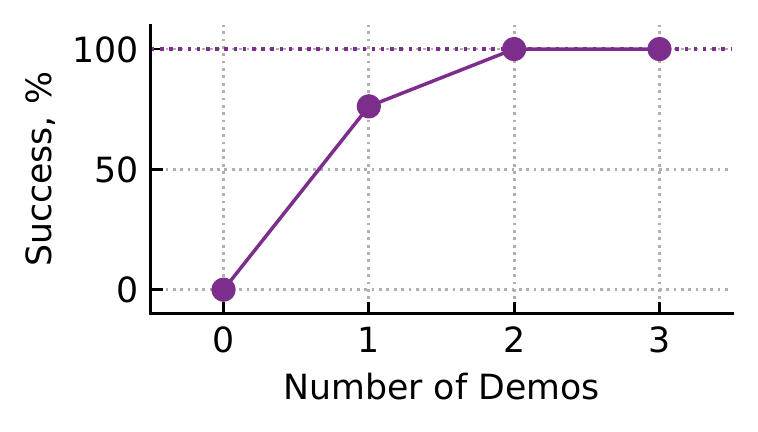}
    \caption{\texttt{Roundabouts1-v0}}
  \end{subfigure}
\caption{
  Adaptation scores of AdaRIP (cf. Section~\ref{sec:adaptive-robust-imitative-planning}) on \texttt{CARNOVEL} tasks that RIP-WCM and RIP-MA (cf. Section~\ref{sec:robust-imitative-planning}) do worst.
  We observe that as the number of online expert demonstrations increases, the success rate improves thanks to online model adaptation.
}
\label{fig:adaptation-scores}
\end{figure*}

\subsubsection{Baselines}
\label{subsub:carnovel-baselines}

We compare RIP against the current state-of-the-art imitation learning methods in the CARLA benchmark~\citep{codevilla2018end,rhinehart2020deep,chen2019learning}.
Apart from DIM and RIP-BCM, discussed in Section~\ref{subsub:nuscenes-baselines}, we also benchmark:

\textbf{Conditional imitation learning}
~\citep[CIL]{codevilla2018end} is a discriminative behavioural cloning method that conditions its predictions on contextual information (e.g., LIDAR) and high-level commands (e.g., turn left, go straight).

\textbf{Learning by cheating}
~\citep[LbC]{chen2019learning} is a method that builds on CIL and uses (cross-modal) distillation of privileged information (e.g., game state, rich, annotated bird-eye-view observations) to a sensorimotor agent.
For reference, we also evaluate the agent who has uses privileged information directly (i.e., teacher), which we term LbC-GT.

\subsubsection{Online Planning Experiments}
\label{subsub:online-planning-experiments}

All the methods are trained on offline expert demonstrations from CARLA \texttt{Town01}.
We perform 10 trials per \texttt{CARNOVEL} task with randomised initial simulator state and the results are reported on Table~\ref{tab:carnovel} and Appendix~\ref{app:experimental-results-on-carnovel}.

Our robust imitative planning (i.e., RIP-WCM and RIP-MA) consistently outperforms the current state-of-the-art imitation learning-based methods in novel, OOD driving scenarios.
In alignment with the experimental results from \texttt{nuScenes} (cf. Section~\ref{sub:nuscenes}), we address questions \textbf{Q4} and \textbf{Q2}, reaching the conclusion that RIP's epistemic uncertainty explicit mechanism for recovery improves its performance under distribution shifts, compared to epistemic uncertainty-unaware methods.
As a result, RIP's recovery score (cf. Section~\ref{subsub:carnovel-metrics}) is higher than the baselines.

\begin{figure}[!h]
  \centering
  \includegraphics[width=\linewidth]{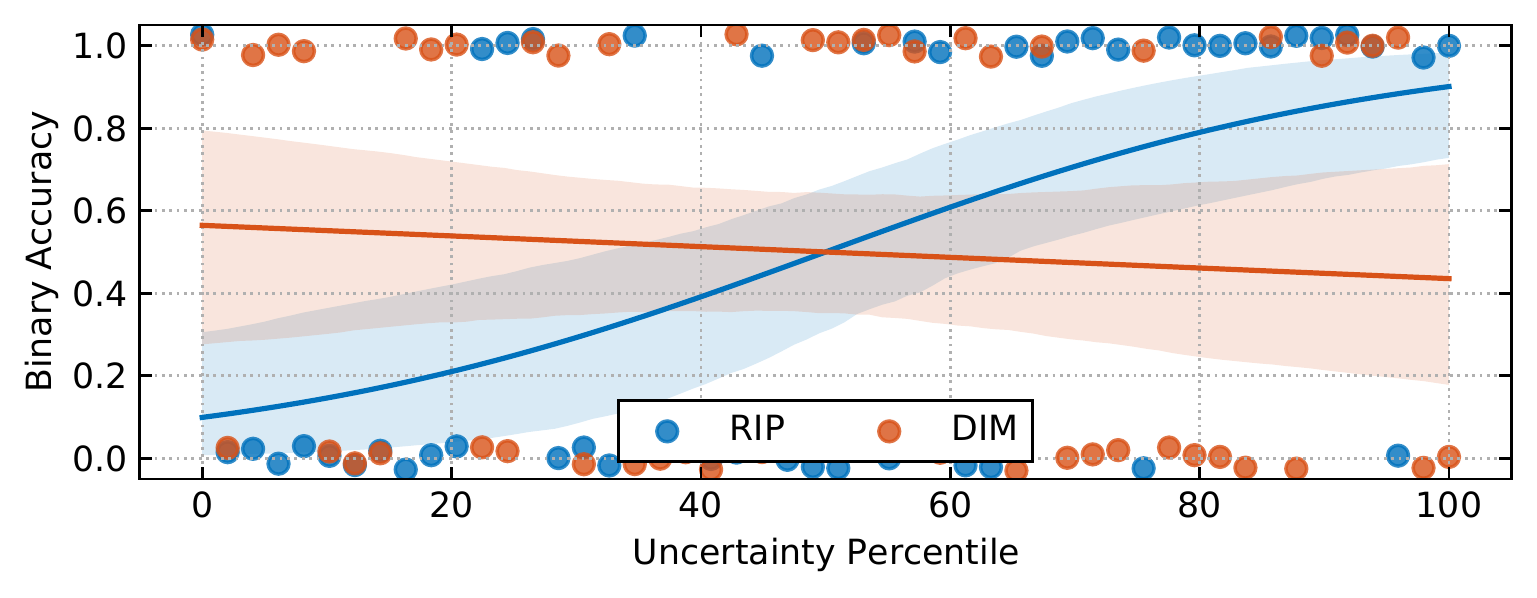}
\caption{
  Uncertainty estimators as indicators of catastrophes on \texttt{CARNOVEL}.
  We collect 50 scenes for each model that led to a crash, record the uncertainty 4 seconds~\citep{taoka1989brake} before the accident and assert if the uncertainties can be used for detection.
  RIP's (ours) predictive variance (\textcolor{matlab-blue}{in blue}, cf. Eqn.~(\ref{eq:var})) serves as a useful detector, while DIM's~\citep{rhinehart2020deep} negative log-likelihood (\textcolor{matlab-orange}{in orange}) cannot be used for detecting catastrophes.
}
\label{fig:detection}
\end{figure}

Towards distribution shift detection and answering questions \textbf{Q1} and \textbf{Q3}, we collect 50 scenes for each method that led to a crash, record the uncertainty 4 seconds~\citep{taoka1989brake} before the accident and assert if the uncertainties can be used for detection.
RIP's (ours) predictive variance (cf. Eqn.~(\ref{eq:var})) serves as a useful detector, while DIM's~\citep{rhinehart2020deep} negative log-likelihood was unable to detect catastrophes.
The results are illustrated on Figure~\ref{fig:detection}.

Despite RIP's improvement over current state-of-the-art methods with $97.5\%$ success rate and $0.26$ infractions per driven kilometre (cf. Table~\ref{tab:carnovel}), the safety-critical nature of the task mandates higher performance. Towards this goal,  we introduce an online adaptation variant of RIP.

\section{Adaptive Robust Imitative Planning}
\label{sec:adaptive-robust-imitative-planning}

\begin{figure*}[t]
  \centering
  \begin{subfigure}[b]{0.24\linewidth}
    \centering
    \begin{tikzpicture}
      \tikzstyle{empty}=[]
      \draw [matlab-red,fill=matlab-red!5] plot [smooth cycle] coordinates {(0,0) (1,1) (3,1) (1.25,0) (2,-1)};
      \draw [matlab-blue,fill=matlab-blue!10]  plot (0.75,0.3) circle (0.35cm);
      \node (A) at (2.5, -0.25) {\small{\color{matlab-red}novel (OOD)}};
      \node (B) at (2.5, 1) {};
      \draw[black,->] (B) -- (A);
      \node (C) at (0.75, 1.5) {\small{\color{matlab-blue}in-distribution}};
      \node (D) at (0.75,0.3) {};
      \draw[black,->] (D) -- (C);
    \end{tikzpicture}
  \caption{Data distribution}
  \label{fig:data-distribution}
  \end{subfigure}
  \begin{subfigure}[b]{0.24\linewidth}
  \centering
  \begin{tikzpicture}
    \tikzstyle{empty}=[]
    \draw [matlab-blue,fill=matlab-blue!10]  plot [smooth cycle] coordinates {(-0.25,-1.25) (-0.25, 1.0) (3.0, 1.5) (3.0, -1.0)};
    \draw [matlab-red,fill=matlab-red!5] plot [smooth cycle] coordinates {(0,0) (1,1) (3,1) (1.25,0) (2,-1)};
  \end{tikzpicture}
  \caption{Domain Randomization}
  \label{fig:domain-randomization}
  \end{subfigure}
  \begin{subfigure}[b]{0.24\linewidth}
  \centering
  \begin{tikzpicture}[scale=0.75]
    \tikzstyle{empty}=[]
    \draw [matlab-red,fill=matlab-red!5] plot [smooth cycle] coordinates {(0,0) (1,1) (3,1) (1.25,0) (2,-1)};
    \draw [matlab-blue,fill=matlab-blue!10]  plot (0.75,0.3) circle (0.35cm);
    \draw [matlab-yellow,fill=matlab-yellow!10] plot coordinates {(0.75,2) (0.75,3) (2.25,3.5) (2.25,1.5)} -- cycle;
    \node (Ain)   at (2.25,2.5)   {};
    \node (Aout)  at (0.75,2.5)   {};
    \node (B)     at (2.5, 0.65)  {};
    \node (C)     at (0.75, 0.25) {};
    \node (ENC)   at (1.5,2.5)   {ENC};
    \path (B) edge [->, bend right, matlab-red]  (Ain);
    \path (Aout) edge [->, bend right, matlab-blue]  (C);
  \end{tikzpicture}
  \caption{Domain adaptation}
  \label{fig:domain-adaptation}
  \end{subfigure}
  \begin{subfigure}[b]{0.24\linewidth}
  \centering
  \begin{tikzpicture}
    \tikzstyle{empty}=[]
    \draw [matlab-red,fill=matlab-red!5] plot [smooth cycle] coordinates {(0,0) (1,1) (3,1) (1.25,0) (2,-1)};
    \draw [matlab-blue,fill=matlab-blue!10] plot [smooth cycle] coordinates {(0.15,0) (0.75,0.75 ) (2.25,0.85) (1,0) (1.25,-0.45)};
    \node (A) at (0.75,0.3) {};
    \node (B) at (1.25,0.8) {};
    \node (C) at (1.25,-0.2) {};
    \node (D) at (0.25,-0.2) {};
    \node (E) at (0.25,0.8) {};
    \draw[matlab-blue,->] (A) -- (B);
    \draw[matlab-blue,->] (A) -- (C);
    \draw[matlab-blue,->] (A) -- (D);
    \draw[matlab-blue,->] (A) -- (E);
  \end{tikzpicture}
  \caption{Online adaptation}
  \label{fig:online-adaptation}
\end{subfigure}
\caption{
  Common approaches to distribution shift, as in (a) there are novel (OOD) points that are outside the support of the training data: (b) domain randomization (e.g.,~\citet{sadeghi2016cad2rl}) covers the data distribution by \emph{exhaustively} sampling configurations from a simulator; (c) domain adaptation (e.g.,~\citet{mcallister2019robustness}) projects (or encodes) the (OOD) points to the in-distribution space and (d) online adaptation (e.g.,~\citet{ross2011dagger}) progressively expands the in-distribution space by incorporating online, external feedback.
}
\label{fig:adaptation-methods}
\end{figure*}
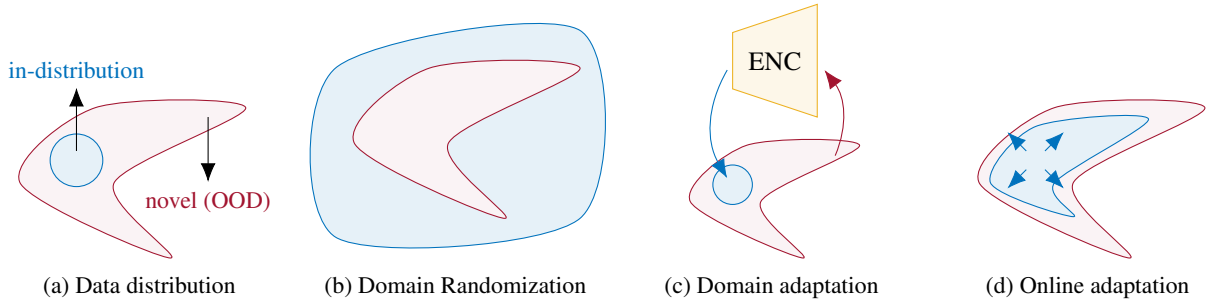

We empirically observe that the quantification of epistemic uncertainty and its use in the RIP objectives is not always sufficient to recover from shifts away from the training distribution (cf. Section~\ref{subsub:online-planning-experiments}).
However, we can use uncertainty estimates to ask the human driver to take back control or default to a safe policy, avoiding potential infractions.
In the former case, the human driver's behaviors can be recorded and used to reduce RIP's epistemic uncertainty via online adaptation.
The epistemic uncertainty is reducible and hence it can be eliminated, provided enough demonstrations.

\begin{algorithm}[h]
  \SetEndCharOfAlgoLine{}
  \SetKwComment{Comment}{// }{}
  \SetKwInOut{Input}{Input}
  \Input{\\\hspace{-3.6em}\small
  \begin{tabular}[t]{l @{\hspace{.25em}} l}%
  $\mathcal{D}$ & Demonstrations \\
  $K$ & Number of models \\
  {\color{matlab-blue}$\mathcal{B}$} & {\color{matlab-blue}Data buffer} \\
  {\color{matlab-blue}$\tau$} & {\color{matlab-blue}Variance threshold} \\
  \end{tabular}\hspace{-0.5em}%
  \begin{tabular}[t]{l @{\hspace{.25em}} l}%
  $\mathbb{I}(a_t \vert s_{t},s_{t+1})$ & Local planner \\
  $q(\by \vert \bx; \btheta)$ & Imitative model \\
  $p(\mathcal{G} \vert \by)$ & Goal likelihood \\
  $p(\btheta)$ & Model prior \\
  \end{tabular}%
  }
  \BlankLine
  \Comment{Approximate model posterior inference, e.g., deep ensemble}
  \For{model index $k=1 \ldots K$}{
    Bootstrap sample dataset $\mathcal{D}_{k} \overset{\text{boot}}{\sim} \mathcal{D}$ \;
    Sample model parameters from prior, $\btheta_{k} \sim p(\btheta)$ \;
    Train ensemble's $k$-component via maximum likelihood estimation (MLE) \texttt{// Eqn.~(\ref{eq:mle})} $\btheta_{k} \leftarrow \argmax_{\btheta} \mathbb{E}_{(\bx, \by) \sim \mathcal{D}_{k}}\left[ \log q(\by \vert \bx ; \btheta) \right]$
  }
  \BlankLine
  \Comment{Online planning}
  $\bx, \mathcal{G} \leftarrow \texttt{env.reset()}$ \;
  \While{not \texttt{done}}{
    Get robust imitative plan \texttt{// Eqn.~(\ref{eq:generic-objective})} $\by^{*} \leftarrow \argmax_{\by} \underset{\btheta}{\oplus} \log q(\by \vert \bx; \btheta) + \log p(\mathcal{G} \vert \by)$
    \BlankLine
    {\color{matlab-blue}
    \Comment{Online adaptation}
    Estimate the predictive variance of the $\by^{*}$ plan's quality under the model posterior \texttt{// Eqn.~(\ref{eq:var})} $u(\by^{*}) = \text{Var}_{p(\btheta \vert \mathcal{D})}\left[ \log q(\by^{*} \vert \bx; \btheta) \right]$  \;
    \If{$u(\by^{*}) > \tau$}{
      $\by^{*} \leftarrow$ Query expert at $\bx$ \;
      $\mathcal{B} \leftarrow \mathcal{B} \cup (\bx, \by^{*})$ \;
      Update model posterior on $\mathcal{B}$ \Comment{with any few-shot adaptation method}
    }
    }
    \BlankLine
    $a_{t} \leftarrow \mathbb{I}(\cdot \vert \by^{*})$ \;
    $\bx, \mathcal{G}, \texttt{done} \leftarrow \texttt{env.step(}a_{t}\texttt{)}$
  }
  \caption{{\color{matlab-blue}Adaptive} Robust Imitative Planning}
  \label{algo:adarip}
\end{algorithm}

We propose an adaptive variant of RIP, called AdaRIP, which uses the epistemic uncertainty estimates to decide when to query the human driver for feedback, which is used to update its parameters \emph{online}, adapting to arbitrary new driving scenarios.
AdaRIP relies on external, online feedback from an expert demonstrator\footnote{AdaRIP is also compatible with other feedback mechanisms, such as expert preferences~\citep{christiano2017deep} or explicit reward functions~\citep{de2019causal}.}, similar to DAgger~\citep{ross2011dagger} and its variants~\citep{zhang2016query, cronrath2018bagger}.
However, unlike this prior work, AdaRIP uses an epistemic uncertainty-aware acquisition mechanism.
AdaRIP's pseudocode is given in Algorithm~\ref{algo:adarip}.

The uncertainty (i.e., variance) threshold, $\tau$, is calibrated on a validation dataset, such that it matches a pre-specified level of false negatives, using a similar analysis to Figure~\ref{fig:detection}.

\vspace{-1em}
\section{Benchmarking Adaptation}
\label{sec:benchmarking-adaptation}

The goal of this section is to provide experimental evidence for answering the following questions:
\textbf{Q5.} Can RIP's epistemic-uncertainty estimation be used for efficiently querying an expert for online feedback (i.e., demonstrations)?
\textbf{Q6.} Does AdaRIP's online adaptation mechanism improve success rate?

We evaluate AdaRIP on \texttt{CARNOVEL} tasks, where the CARLA autopilot~\citep{dosovitskiy2017carla} is queried for demonstrations online when the predictive variance (cf. Eqn.~(\ref{eq:var})) exceeds a threshold, chosen according to RIP's detection score, (cf. Figure~\ref{fig:detection}).
We measure performance according to the:

\textbf{Adaptation score.}
The improvement in success rate as a function of number of online expert demonstrations is used to measure a method's ability to adapt efficiently online.
We refer to this metric as \emph{adaptation score}.
A method that can adapt online should have a positive adaptation score.

AdaRIP's performance on the most challenging \texttt{CARNOVEL} tasks is summarised in Figure~\ref{fig:adaptation-scores}, where, as expected, the success rate improves as the number of online demonstrations increases.
Qualitative examples are illustrated in Appendix~\ref{app:adarip}.

Although AdaRIP can adapt to any distribution shift, it is prone to catastrophic forgetting and sample-inefficiency, as many online methods~\citep{french1999catastrophic}.
In this paper, we only demonstrate AdaRIP's efficacy to adapt under distribution shifts and do not address either of these limitations.
Future work lies in providing a practical, sample-efficient algorithm to be used in conjunction with the AdaRIP framework.
Methods for efficient (e.g., few-shot or zero-shot) and safe adaptation~\citep{finn2017model, zhou2019watch} are orthogonal to AdaRIP and hence any improvement in these fields could be directly used for AdaRIP.

\section{Related Work}
\label{sec:related-work}

\textbf{Imitation learning.}
~Learning from expert demonstrations (i.e., imitation learning~\citep[IL]{widrow1964pattern, pomerleau1989alvinn}) is an attractive framework for sequential decision-making in safety-critical domains such as autonomous driving, where trial and error learning has little to no safety guarantees during training.
A plethora of expert driving demonstrations has been used for IL~\citep{caesar2019nuscenes, sun2019scalability, lyft2019} since a model mimicking expert demonstrations can simply learn to stay in ``safe'', expert-like parts of the state space and no explicit reward function need be specified.

On the one hand, behavioural cloning approaches~\citep{liang2018cirl, sauer2018conditional, li2018rethinking, codevilla2018end, codevilla2019exploring, chen2019learning} fit command-conditioned discriminative sequential models to expert demonstrations, which are used in deployment to produce expert-like trajectories.
On the other hand,~\citet{rhinehart2020deep} proposed command-\emph{unconditioned} expert trajectory density models which are used for planning trajectories that both satisfy the goal constraints and are likely under the expert model.
However, both of these approaches fit point-estimates to their parameters, thus do not quantify their model (\emph{epistemic}) uncertainty, as explained next.
This is especially problematic when estimating what an expert would or would not do in \emph{unfamiliar}, OOD scenes.
In contrast, our methods, RIP and AdaRIP, does quantify epistemic uncertainty in order to both improve planning performance and triage situations in which an expert should intervene.

\textbf{Novelty detection \& epistemic uncertainty.}
~A principled means to capture epistemic uncertainty is with Bayesian inference to compute the predictive distribution.
However, evaluating the posterior $p(\btheta \vert \mathcal{D})$ with exact inference is intractable for non-trivial models~\citep{neal2012bayesian}.
Approximate inference methods~\citep{graves2011practical, blundell2015weight, gal2016dropout, hernandez2015probabilistic} have been introduced that can efficiently capture epistemic uncertainty.
One approximation for epistemic uncertainty in deep models is model ensembles \citep{lakshminarayanan2017simple,chua2018deep}. Prior work by \citet{kahn2017uncertainty} and \citet{kenton2019generalizing} use ensembles of deep models to detect and avoid catastrophic actions in navigation tasks, although they can not recover from or adapt to distribution shifts.
Our epistemic uncertainty-aware planning objective, RIP, instead, managed to recover from some distribution shifts, as shown experimentally in Section~\ref{sec:benchmarking-robustness-to-novelty}.

\textbf{Coping with distribution shift.}
~Strategies to cope with distribution shift include (a) domain randomization; (b) domain adaptation and (c) online adaptation.
\emph{Domain randomization} assumes access to a simulator and \emph{exhaustively} searches for configurations that cover all the data distribution support in order to eliminate OOD scenes, as illustrated in Figure~\ref{fig:domain-randomization}.
This approach has been successfully used in simple robotic tasks~\citep{sadeghi2016cad2rl, openai2018learning, akkaya2019solving} but it is impractical for use in large, real-world tasks, such as AD.
\emph{Domain adaptation} and \emph{bisimulation}~\citep{castro2010using}, depicted in Figure~\ref{fig:domain-adaptation}, tackle OOD points by projecting them back to in-distribution points, that are ``close'' to training points according to some metric.
Despite its success in simple visual tasks~\citep{mcallister2019robustness}, it has no guarantees under arbitrary distribution shifts.
In contrast, \emph{online learning methods}~\citep{cesa2006prediction, ross2011dagger, zhang2016query, cronrath2018bagger} have no-regret guarantees and, provided frequent expert supervision, they asymptotically cover the whole data distribution's support, adaptive to any distribution shift, as shown in Figure~\ref{fig:online-adaptation}. In order to continually cope with distribution shift, a learner must receive interactive feedback~\citep{ross2011dagger}, however, the frequency of this costly feedback should be minimised. Our epistemic-uncertainty-aware method, Robust Imitative Planning can cope with some OOD events, thereby reducing the system's dependency on expert feedback, and can use this uncertainty to decide when it cannot cope--when the expert must intervene.

\textbf{Current benchmarks.}
~We are interested in the control problem, where AD agents get deployed in reactive environments and make sequential decisions.
The CARLA Challenge~\citep{carlachallenge, dosovitskiy2017carla, codevilla2019exploring} is an open-source benchmark for control in AD.
It is based on 10 traffic scenarios from the NHTSA pre-crash typology~\citep{nhtsa2007pre} to inject challenging driving situations into traffic patterns encountered by AD agents.
The methods are only assessed in terms of their generalization to weather conditions, the initial state of the simulation (e.g., the start and goal locations, and the random seed of other agents.) and the traffic density (i.e., empty town, regular traffic and dense traffic).

Despite these challenging scenarios selected in the CARLA Challenge, the agents are allowed to train on the same scenarios in which they evaluated, and so \emph{the robustness to distributional shift is not assessed}.
Consequently, both \citet{chen2019learning} and \citet{rhinehart2020deep} manage to solve the CARLA Challenge with almost $100\%$ success rate, when trained in \texttt{Town01} and tested in \texttt{Town02}.
However, both methods score \emph{almost} $0\%$ when evaluated in \texttt{Roundabouts} due to the presence of OOD road morphologies, as discussed in Section~\ref{subsub:online-planning-experiments}.

\section{Summary and Conclusions}
\label{sec:summary-and-conclusions}

To summarise, in this paper, we studied autonomous driving agents in out-of-training distribution tasks (i.e. under distribution shifts).
We introduced an epistemic uncertainty-aware planning method, called robust imitative planning (RIP), which can detect and recover from distribution shifts, as shown experimentally in a real prediction task, \texttt{nuScenes}, and a driving simulator, CARLA.
We presented an adaptive variant (AdaRIP) which uses RIP's epistemic uncertainty estimates to efficiently query the expert for online feedback and adapt its model parameters online.
We also introduced and open-sourced an autonomous car novel-scene benchmark, termed \texttt{CARNOVEL}, to assess the robustness of driving agents to a suite of OOD tasks.

\vspace{-1em}

\section*{Acknowledgements}
\label{sec:acknowledgements}

This work was supported by the UK EPSRC CDT in Autonomous Intelligent Machines and Systems (grant reference EP/L015897/1).
This project has received funding from the Office of Naval Research, the DARPA Assured Autonomy Program, and ARL DCIST CRA W911NF-17-2-0181, Microsoft Azure and Intel AI Labs.

\bibliography{references}
\bibliographystyle{include/icml2020}

\clearpage
\appendix
\onecolumn

\section{\texttt{CARNOVEL}: Suite of Tasks Under Distribution Shift}
\label{app:carnovel}

\begin{figure}[h]
  \centering
  \foreach \task in {%
    AbnormalTurns0-v0,%
    AbnormalTurns1-v0,%
    AbnormalTurns2-v0%
    }{\begin{subfigure}[b]{0.32\linewidth}
      \centering
      \begin{subfigure}[b]{0.48\linewidth}
        \centering
        \includegraphics[width=\textwidth]{fig/CARNOVEL/spawn_points/\task.png}
      \end{subfigure}
      \begin{subfigure}[b]{0.48\linewidth}
        \centering
        \includegraphics[width=\textwidth]{fig/CARNOVEL/tasks/\task.png}
      \end{subfigure}
    \caption{\texttt{\task}}
    \end{subfigure}}
  \\~\\
  \foreach \task in {%
    AbnormalTurns3-v0,%
    AbnormalTurns4-v0,%
    AbnormalTurns5-v0%
    }{\begin{subfigure}[b]{0.32\linewidth}
      \centering
      \begin{subfigure}[b]{0.48\linewidth}
        \centering
        \includegraphics[width=\textwidth]{fig/CARNOVEL/spawn_points/\task.png}
      \end{subfigure}
      \begin{subfigure}[b]{0.48\linewidth}
        \centering
        \includegraphics[width=\textwidth]{fig/CARNOVEL/tasks/\task.png}
      \end{subfigure}
    \caption{\texttt{\task}}
    \end{subfigure}}
  \\~\\
  \foreach \task in {%
    AbnormalTurns6-v0,%
    BusyTown0-v0,%
    BusyTown1-v0%
    }{\begin{subfigure}[b]{0.32\linewidth}
      \centering
      \begin{subfigure}[b]{0.48\linewidth}
        \centering
        \includegraphics[width=\textwidth]{fig/CARNOVEL/spawn_points/\task.png}
      \end{subfigure}
      \begin{subfigure}[b]{0.48\linewidth}
        \centering
        \includegraphics[width=\textwidth]{fig/CARNOVEL/tasks/\task.png}
      \end{subfigure}
    \caption{\texttt{\task}}
    \end{subfigure}}
  \\~\\
  \foreach \task in {%
    BusyTown2-v0,%
    BusyTown3-v0,%
    BusyTown4-v0%
    }{\begin{subfigure}[b]{0.32\linewidth}
      \centering
      \begin{subfigure}[b]{0.48\linewidth}
        \centering
        \includegraphics[width=\textwidth]{fig/CARNOVEL/spawn_points/\task.png}
      \end{subfigure}
      \begin{subfigure}[b]{0.48\linewidth}
        \centering
        \includegraphics[width=\textwidth]{fig/CARNOVEL/tasks/\task.png}
      \end{subfigure}
    \caption{\texttt{\task}}
    \end{subfigure}}
  \\~\\
  \foreach \task in {%
    BusyTown5-v0,%
    BusyTown6-v0,%
    BusyTown7-v0%
    }{\begin{subfigure}[b]{0.32\linewidth}
      \centering
      \begin{subfigure}[b]{0.48\linewidth}
        \centering
        \includegraphics[width=\textwidth]{fig/CARNOVEL/spawn_points/\task.png}
      \end{subfigure}
      \begin{subfigure}[b]{0.48\linewidth}
        \centering
        \includegraphics[width=\textwidth]{fig/CARNOVEL/tasks/\task.png}
      \end{subfigure}
    \caption{\texttt{\task}}
    \end{subfigure}}
\end{figure}
\begin{figure}[h]\ContinuedFloat
  \centering
  \foreach \task in {%
    BusyTown8-v0,%
    BusyTown9-v0,%
    BusyTown10-v0%
    }{\begin{subfigure}[b]{0.32\linewidth}
      \centering
      \begin{subfigure}[b]{0.48\linewidth}
        \centering
        \includegraphics[width=\textwidth]{fig/CARNOVEL/spawn_points/\task.png}
      \end{subfigure}
      \begin{subfigure}[b]{0.48\linewidth}
        \centering
        \includegraphics[width=\textwidth]{fig/CARNOVEL/tasks/\task.png}
      \end{subfigure}
    \caption{\texttt{\task}}
    \end{subfigure}}
  \\~\\
  \foreach \task in {%
    Hills0-v0,%
    Hills1-v0,%
    Hills2-v0%
    }{\begin{subfigure}[b]{0.32\linewidth}
      \centering
      \begin{subfigure}[b]{0.48\linewidth}
        \centering
        \includegraphics[width=\textwidth]{fig/CARNOVEL/spawn_points/\task.png}
      \end{subfigure}
      \begin{subfigure}[b]{0.48\linewidth}
        \centering
        \includegraphics[width=\textwidth]{fig/CARNOVEL/tasks/\task.png}
      \end{subfigure}
    \caption{\texttt{\task}}
    \end{subfigure}}
  \\~\\
  \foreach \task in {%
    Hills3-v0,%
    Roundabouts0-v0,%
    Roundabouts1-v0%
    }{\begin{subfigure}[b]{0.32\linewidth}
      \centering
      \begin{subfigure}[b]{0.48\linewidth}
        \centering
        \includegraphics[width=\textwidth]{fig/CARNOVEL/spawn_points/\task.png}
      \end{subfigure}
      \begin{subfigure}[b]{0.48\linewidth}
        \centering
        \includegraphics[width=\textwidth]{fig/CARNOVEL/tasks/\task.png}
      \end{subfigure}
    \caption{\texttt{\task}}
    \end{subfigure}}
  \\~\\
  \foreach \task in {%
    Roundabouts2-v0,%
    Roundabouts3-v0,%
    Roundabouts4-v0%
    }{\begin{subfigure}[b]{0.32\linewidth}
      \centering
      \begin{subfigure}[b]{0.48\linewidth}
        \centering
        \includegraphics[width=\textwidth]{fig/CARNOVEL/spawn_points/\task.png}
      \end{subfigure}
      \begin{subfigure}[b]{0.48\linewidth}
        \centering
        \includegraphics[width=\textwidth]{fig/CARNOVEL/tasks/\task.png}
      \end{subfigure}
    \caption{\texttt{\task}}
    \end{subfigure}}
  \label{fig:CARNOVEL-tasks}
\end{figure}

\clearpage

\section{Experimental Results on \texttt{CARNOVEL}}
\label{app:experimental-results-on-carnovel}

\begin{table*}[h]
  \centering
  \caption{
    We evaluate different autonomous driving methods in terms of their robustness to distribution shifts, in our new benchmark, \texttt{CARNOVEL}.
    All methods are trained on CARLA \texttt{Town01} using imitation learning on expert demonstrations from the autopilot~\citep{dosovitskiy2017carla}.
    A ``$\dagger$'' indicates methods that use first-person camera view, as in~\citep{chen2019learning}, a ``$\clubsuit$'' methods that use LIDAR observation, as in~\citep{rhinehart2020deep} and a ``$\diamondsuit$'' methods that use the ground truth game engine state, as in~\citep{chen2019learning}.
    A ``$\bigstar$'' indicates that we used the reference implementation from the original paper, otherwise we used our implementation.
    For all the scenes we chose pairs of start-destination locations and ran $10$ trials with randomized initial simulator state for each pair.
    \textcolor{black!50}{Standard errors} are in gray (via bootstrap sampling).
    The \textbf{outperforming} method is in bold.
  }
  \label{tab:results}
  \resizebox{\linewidth}{!}{
  \begin{tabular}{llll|lll}
    \toprule
                                                &
    \multicolumn{3}{c}{\texttt{AbnormalTurns}}  &
    \multicolumn{3}{c}{\texttt{BusyTown}}       \\
    \cline{2-3} \cline{4-5} \cline{6-7} \\
                                    &
    Success $\uparrow$              &
    Infra/km $\downarrow$           &
    Distance $\uparrow$             &
    Success $\uparrow$              &
    Infra/km $\downarrow$           &
    Distance $\uparrow$             \\
    \textbf{Methods}                &
    ($7 \times 10$ scenes, $\%$)    &
    ($\times 1e{-3}$)                 &
    (m)                             &
    ($11 \times 10$ scenes, $\%$)     &
    ($\times 1e{-3}$)                 &
    (m)                             \\
    \midrule
    CIL$^{\clubsuit \bigstar}$~\citep{codevilla2018end} &
      $65.71{\color{black!50}\pm07.37}$ &
      $07.04{\color{black!50}\pm05.07}$ &
      $128{\color{black!50}\pm020}$ &
      $05.45{\color{black!50}\pm06.35}$ &
      $11.49{\color{black!50}\pm03.66}$ &
      $217{\color{black!50}\pm033}$ \\
    LbC$^{\dagger \bigstar}$~\citep{chen2019learning} &
      $00.00{\color{black!50}\pm00.00}$ &
      $05.81{\color{black!50}\pm00.58}$ &
      $208{\color{black!50}\pm004}$ &
      $20.00{\color{black!50}\pm13.48}$ &
      $03.96{\color{black!50}\pm00.15}$ &
      $374{\color{black!50}\pm016}$ \\
    LbC-GT$^{\diamondsuit \bigstar}$~\citep{chen2019learning} &
      $02.86{\color{black!50}\pm06.39}$ &
      \bftab{03.68}${\color{black!50}\pm00.34}$ &
      $217{\color{black!50}\pm033}$ &
      $65.45{\color{black!50}\pm07.60}$ &
      $02.59{\color{black!50}\pm00.02}$ &
      $400{\color{black!50}\pm006}$ \\
    \midrule
    DIM$^{\clubsuit}$~\citep{rhinehart2020deep} &
      $74.28{\color{black!50}\pm11.26}$ &
      $05.56{\color{black!50}\pm04.06}$ &
      $108{\color{black!50}\pm017}$ &
      $47.13{\color{black!50}\pm14.54}$ &
      $08.47{\color{black!50}\pm05.22}$ &
      $175{\color{black!50}\pm026}$ \\
    RIP-BCM$^{\clubsuit}$ (baseline, cf. Table~\ref{tab:methods}) &
      $68.57{\color{black!50}\pm09.03}$ &
      $07.93{\color{black!50}\pm03.73}$ &
      $096{\color{black!50}\pm017}$ &
      $50.90{\color{black!50}\pm20.64}$ &
      $03.74{\color{black!50}\pm05.52}$ &
      $175{\color{black!50}\pm031}$ \\
    \midrule
    \rowcolor{ourmethod}
    RIP-MA$^{\clubsuit}$ (ours, cf. Section~\ref{subsubsec:model-averaging}) &
      \bftab{84.28}${\color{black!50}\pm14.20}$ &
      $07.86{\color{black!50}\pm05.70}$ &
      $102{\color{black!50}\pm015}$ &
      $\mathbf{64.54{\color{black!50}\pm23.25}}$ &
      $05.86{\color{black!50}\pm03.99}$ &
      $170{\color{black!50}\pm033}$ \\
    \rowcolor{ourmethod}
    RIP-WCM$^{\clubsuit}$ (ours, cf. Section~\ref{subsubsec:worst-case-model}) &
      \bftab{87.14}${\color{black!50}\pm14.20}$ &
      \bftab{04.91}${\color{black!50}\pm03.60}$ &
      $102{\color{black!50}\pm021}$ &
      \bftab{62.72}${\color{black!50}\pm05.16}$ &
      \bftab{03.17}${\color{black!50}\pm02.04}$ &
      $167{\color{black!50}\pm021}$ \\
    \bottomrule
  \end{tabular}}

  \vspace{0.5em}

  \resizebox{\linewidth}{!}{
  \begin{tabular}{llll|lll}
    \toprule
                                              &
    \multicolumn{3}{c}{\texttt{Hills}}        &
    \multicolumn{3}{c}{\texttt{Roundabouts}}  \\
    \cline{2-3} \cline{4-5} \cline{6-7} \\
    &
    Success $\uparrow$              &
    Infra/km $\downarrow$           &
    Distance $\uparrow$             &
    Success $\uparrow$              &
    Infra/km $\downarrow$           &
    Distance $\uparrow$             \\
    \textbf{Methods}                &
    ($4 \times 10$ scenes, $\%$)    &
    ($\times 1e{-3}$)                 &
    (m)                             &
    ($5 \times 10$ scenes, $\%$)    &
    ($\times 1e{-3}$)                 &
    (m)                            \\
    \midrule
    CIL$^{\clubsuit \bigstar}$~\citep{codevilla2018end} &
      $60.00{\color{black!50}\pm29.34}$ &
      $04.74{\color{black!50}\pm03.02}$ &
      $219{\color{black!50}\pm034}$ &
      $20.00{\color{black!50}\pm00.00}$ &
      \bftab{03.60}${\color{black!50}\pm03.23}$ &
      $269{\color{black!50}\pm021}$ \\
    LbC$^{\dagger \bigstar}$~\citep{chen2019learning} &
      $50.00{\color{black!50}\pm00.00}$ &
      $01.61{\color{black!50}\pm00.15}$ &
      $541{\color{black!50}\pm101}$ &
      $08.00{\color{black!50}\pm10.95}$ &
      $03.70{\color{black!50}\pm00.72}$ &
      $323{\color{black!50}\pm043}$ \\
    LbC-GT$^{\diamondsuit \bigstar}$~\citep{chen2019learning} &
      $05.00{\color{black!50}\pm11.18}$ &
      $03.36{\color{black!50}\pm00.26}$ &
      $312{\color{black!50}\pm020}$ &
      $00.00{\color{black!50}\pm00.00}$ &
      $06.47{\color{black!50}\pm00.99}$ &
      $123{\color{black!50}\pm018}$ \\
    \midrule
    DIM$^{\clubsuit}$~\citep{rhinehart2020deep} &
      $70.00{\color{black!50}\pm10.54}$ &
      $06.87{\color{black!50}\pm04.09}$ &
      $195{\color{black!50}\pm012}$ &
      $20.00{\color{black!50}\pm09.42}$ &
      $06.19{\color{black!50}\pm04.73}$ &
      $240{\color{black!50}\pm044}$ \\
    RIP-BCM$^{\clubsuit}$ (baseline, cf. Table~\ref{tab:methods}) &
      $75.00{\color{black!50}\pm00.00}$ &
      $05.49{\color{black!50}\pm04.03}$ &
      $191{\color{black!50}\pm013}$ &
      $06.00{\color{black!50}\pm09.66}$ &
      $06.78{\color{black!50}\pm07.05}$ &
      $251{\color{black!50}\pm027}$ \\
    \midrule
    \rowcolor{ourmethod}
    RIP-MA$^{\clubsuit}$ (ours, cf. Section~\ref{subsubsec:model-averaging}) &
      \bftab{97.50}${\color{black!50}\pm07.90}$ &
      \bftab{00.26}${\color{black!50}\pm00.54}$ &
      $196{\color{black!50}\pm013}$ &
      \bftab{38.00}${\color{black!50}\pm06.32}$ &
      $05.48{\color{black!50}\pm05.56}$ &
      $271{\color{black!50}\pm047}$ \\
    \rowcolor{ourmethod}
    RIP-WCM$^{\clubsuit}$ (ours, cf. Section~\ref{subsubsec:worst-case-model}) &
      \bftab{87.50}${\color{black!50}\pm13.17}$ &
      \bftab{01.83}${\color{black!50}\pm01.73}$ &
      $191{\color{black!50}\pm006}$ &
      \bftab{42.00}${\color{black!50}\pm06.32}$ &
      $04.32{\color{black!50}\pm01.91}$ &
      $217{\color{black!50}\pm030}$ \\
    \bottomrule
  \end{tabular}}
\end{table*}

\clearpage

\section{AdaRIP Examples}
\label{app:adarip}

\begin{figure}[ht]
  \centering
  \begin{subfigure}[b]{0.5\linewidth}
    \centering
    \caption*{(Normalized) Uncertainty}
    \includegraphics[width=\linewidth]{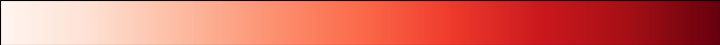}
  \end{subfigure}
  \\~\\
  \begin{subfigure}[b]{0.25\linewidth}
    \centering
    \includegraphics[width=\linewidth]{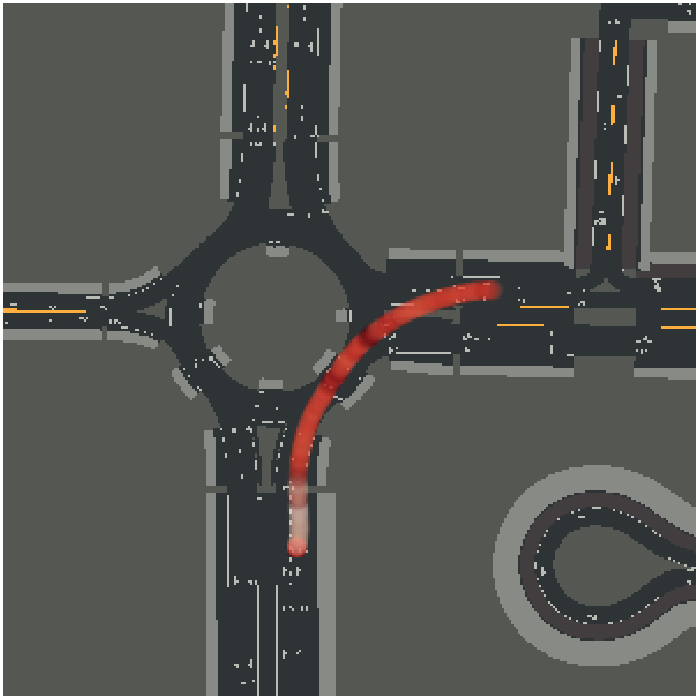}
  \end{subfigure}
  \begin{subfigure}[b]{0.25\linewidth}
    \centering
    \includegraphics[width=\linewidth]{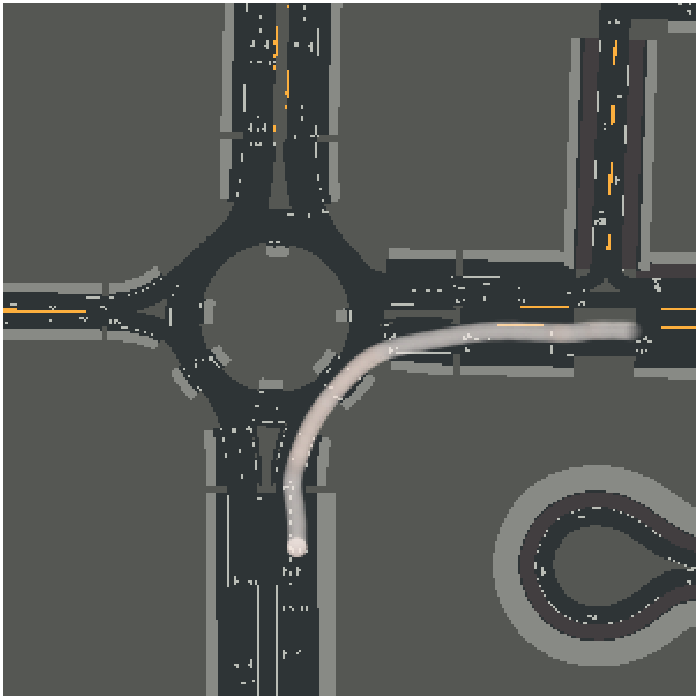}
  \end{subfigure}
  \\~\\
  \begin{subfigure}[b]{0.25\linewidth}
    \centering
    \includegraphics[width=\linewidth]{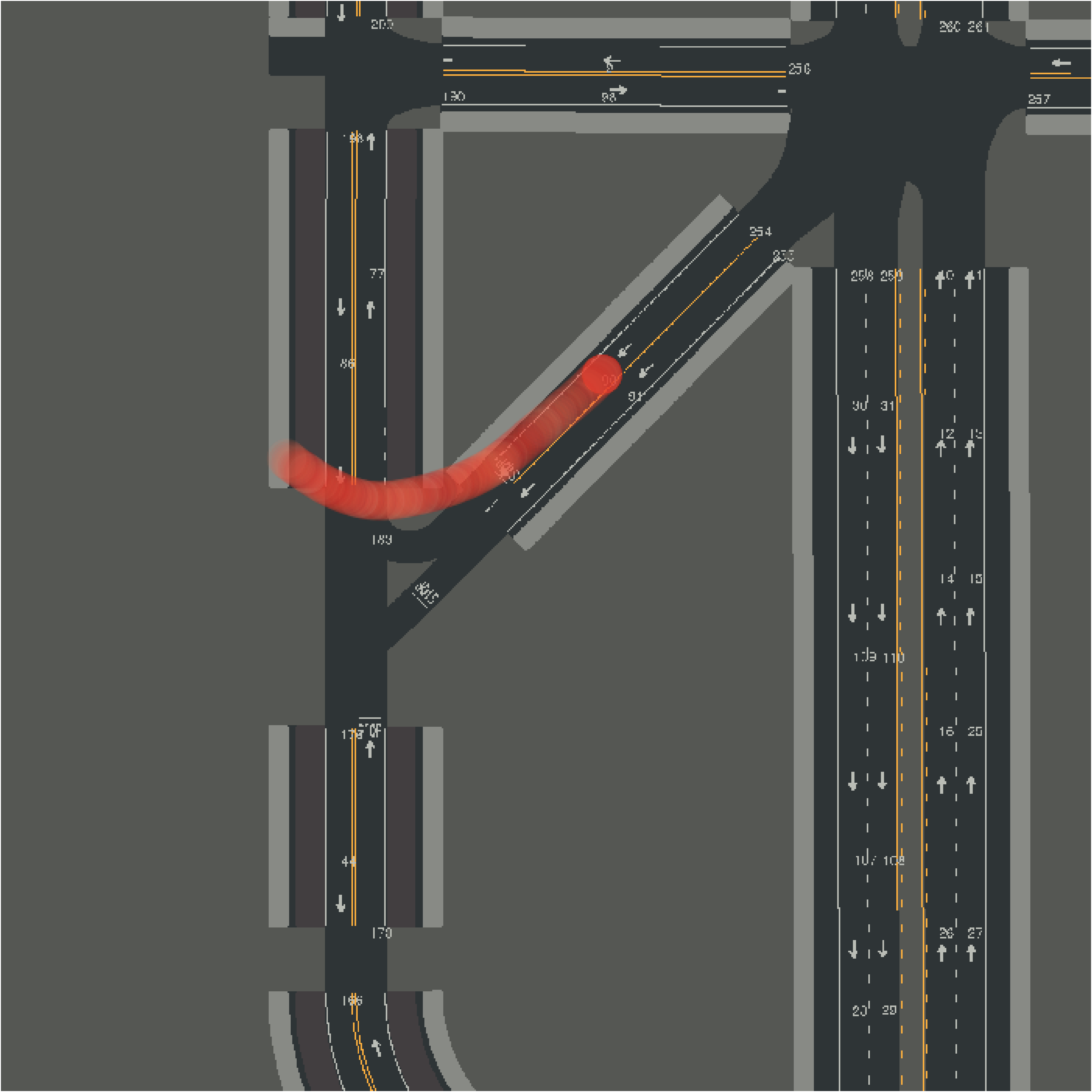}
  \end{subfigure}
  \begin{subfigure}[b]{0.25\linewidth}
    \centering
    \includegraphics[width=\linewidth]{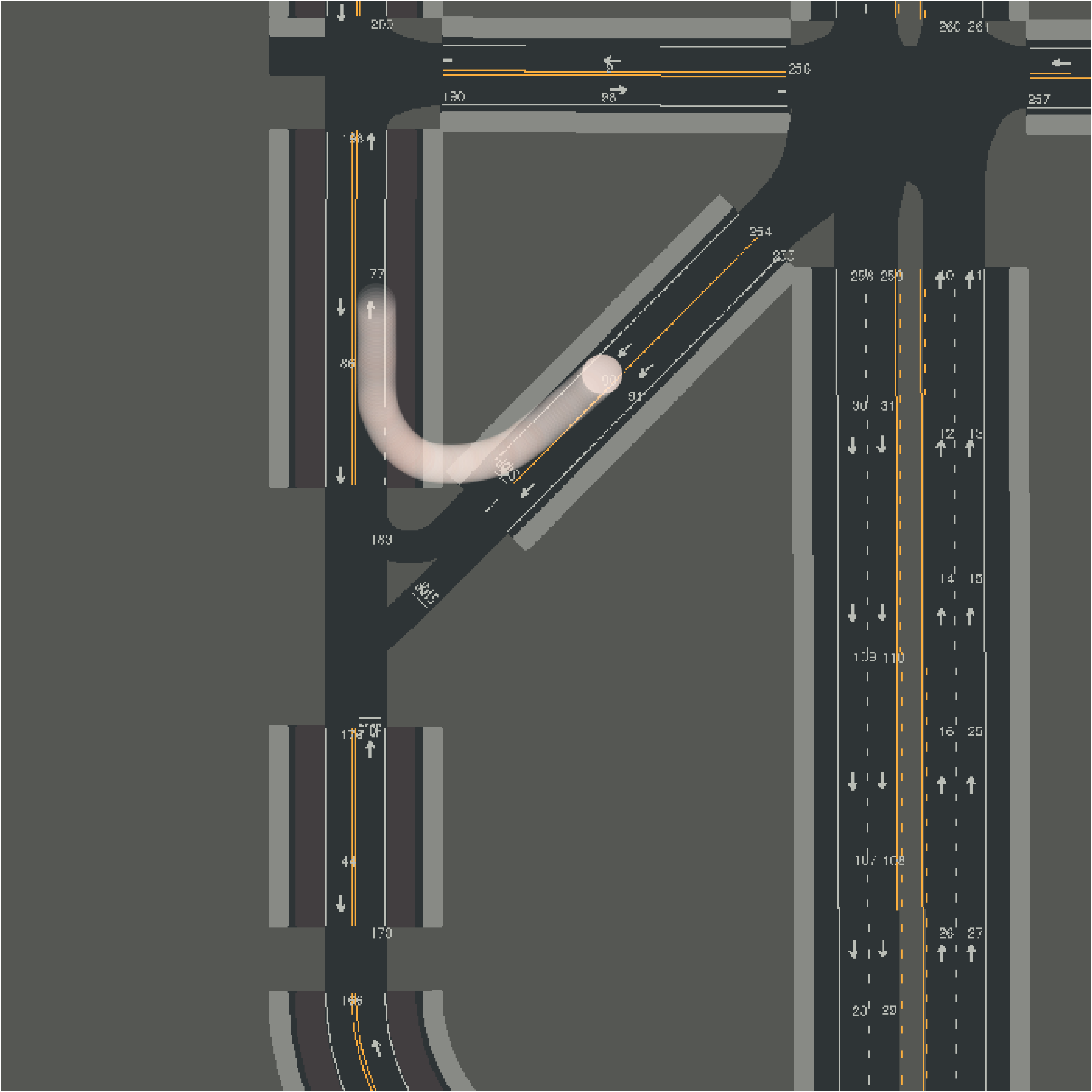}
  \end{subfigure}
  \\~\\
  \begin{subfigure}[b]{0.25\linewidth}
    \centering
    \includegraphics[width=\linewidth]{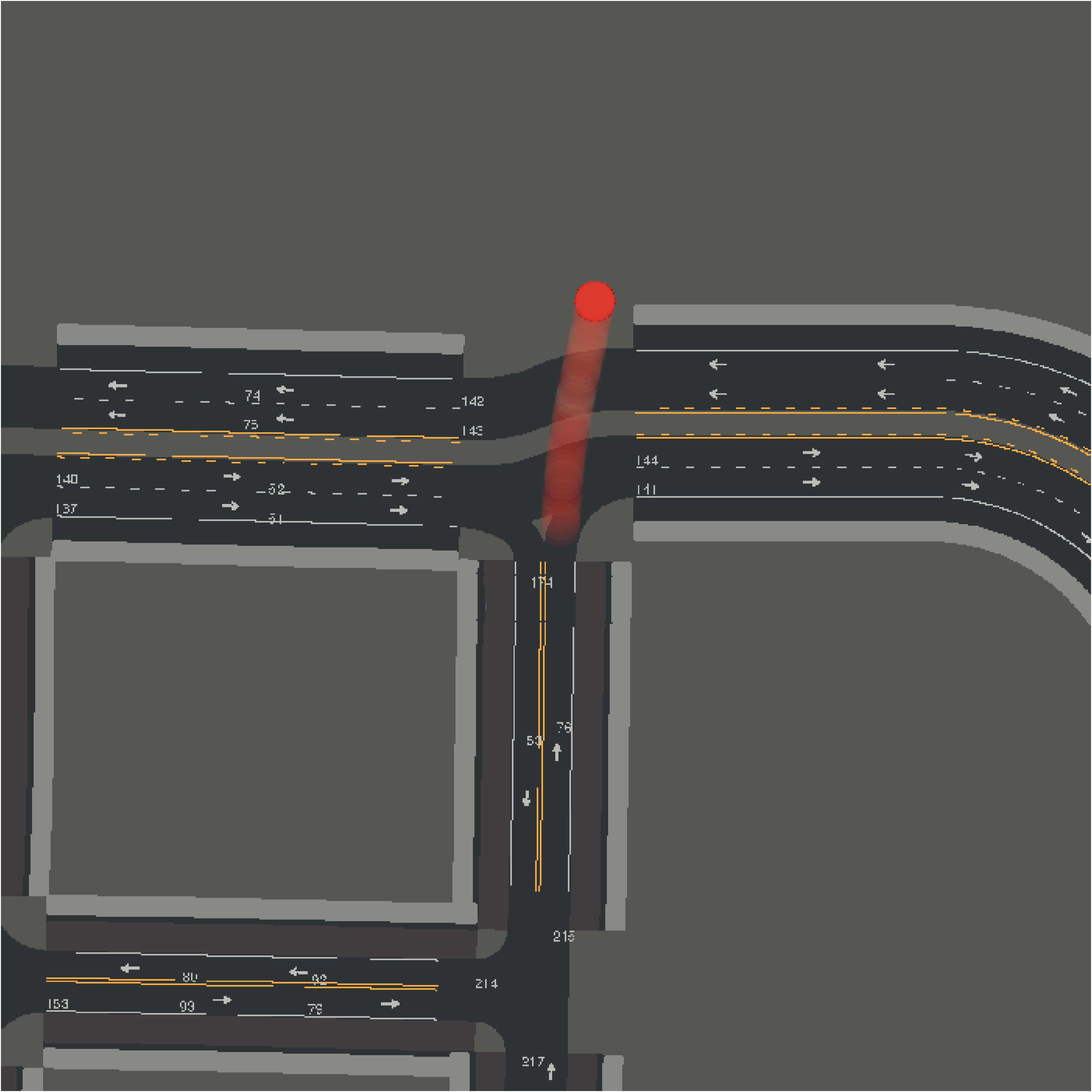}
    \caption{RIP}
    \label{fig:extra-rip}
  \end{subfigure}
  \begin{subfigure}[b]{0.25\linewidth}
    \centering
    \includegraphics[width=\linewidth]{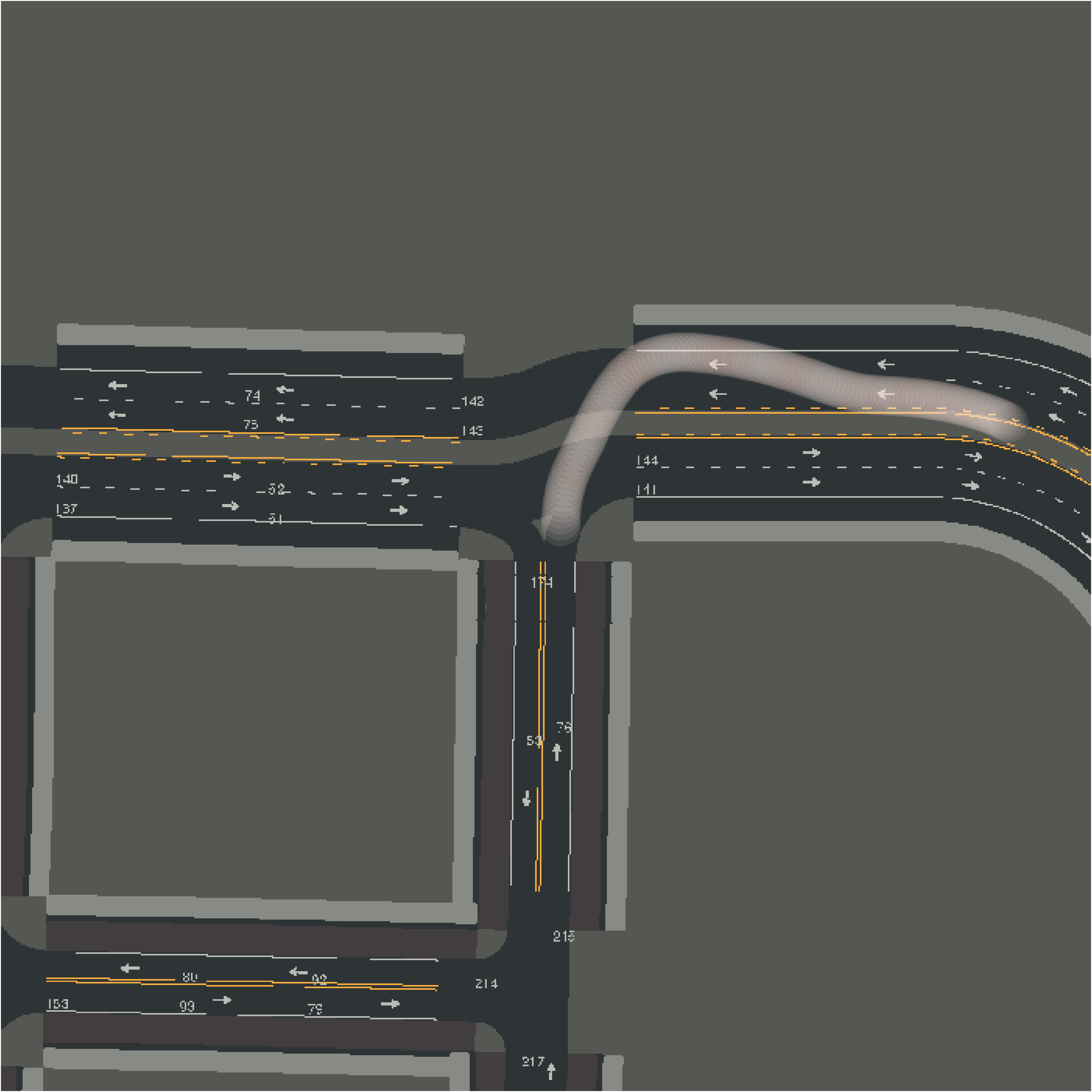}
    \caption{AdaRIP}
    \label{fig:extra-adarip}
  \end{subfigure}
\caption{
  Examples where the non-adaptive method (a) fails to recover from a distribution shift, despite it being able to detect it.
  The adaptive method (b) queries the human driver when uncertain (dark red), then uses the online demonstrations for updating its model, resulting into confident (light red, white) and safe trajectories.
}
\label{fig:adarip-extra}
\end{figure}

\clearpage

\section{Online Planning with a Trajectory Library}
\label{app:onlin-planning-with-a-trajectory-library}

\begin{figure}[h]
  \centering
  \begin{subfigure}[l]{0.24\linewidth}
    \includegraphics[width=\textwidth]{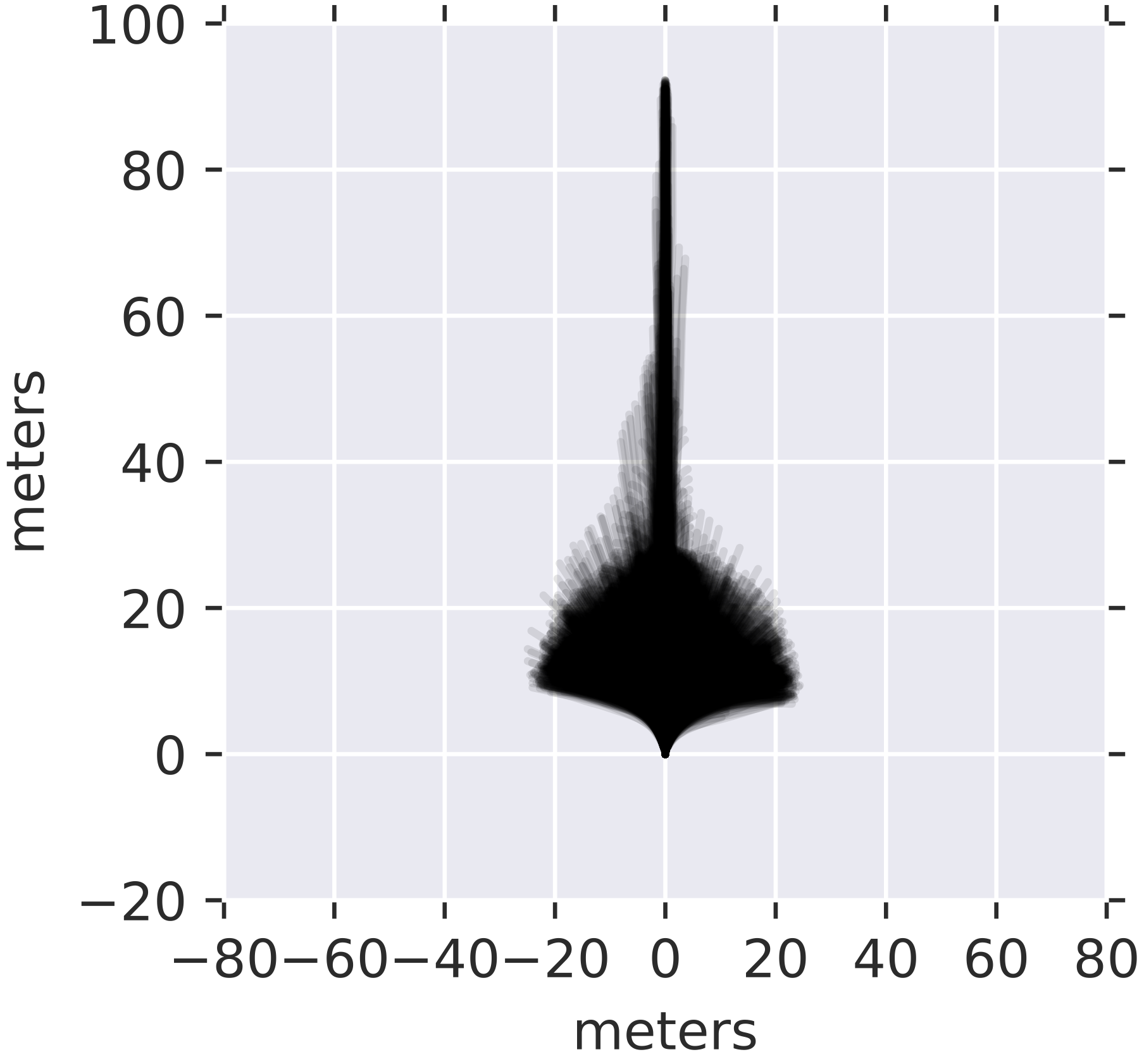}
    \caption{Trajectories}
  \end{subfigure}
  \begin{subfigure}[l]{0.24\linewidth}
    \includegraphics[width=\textwidth]{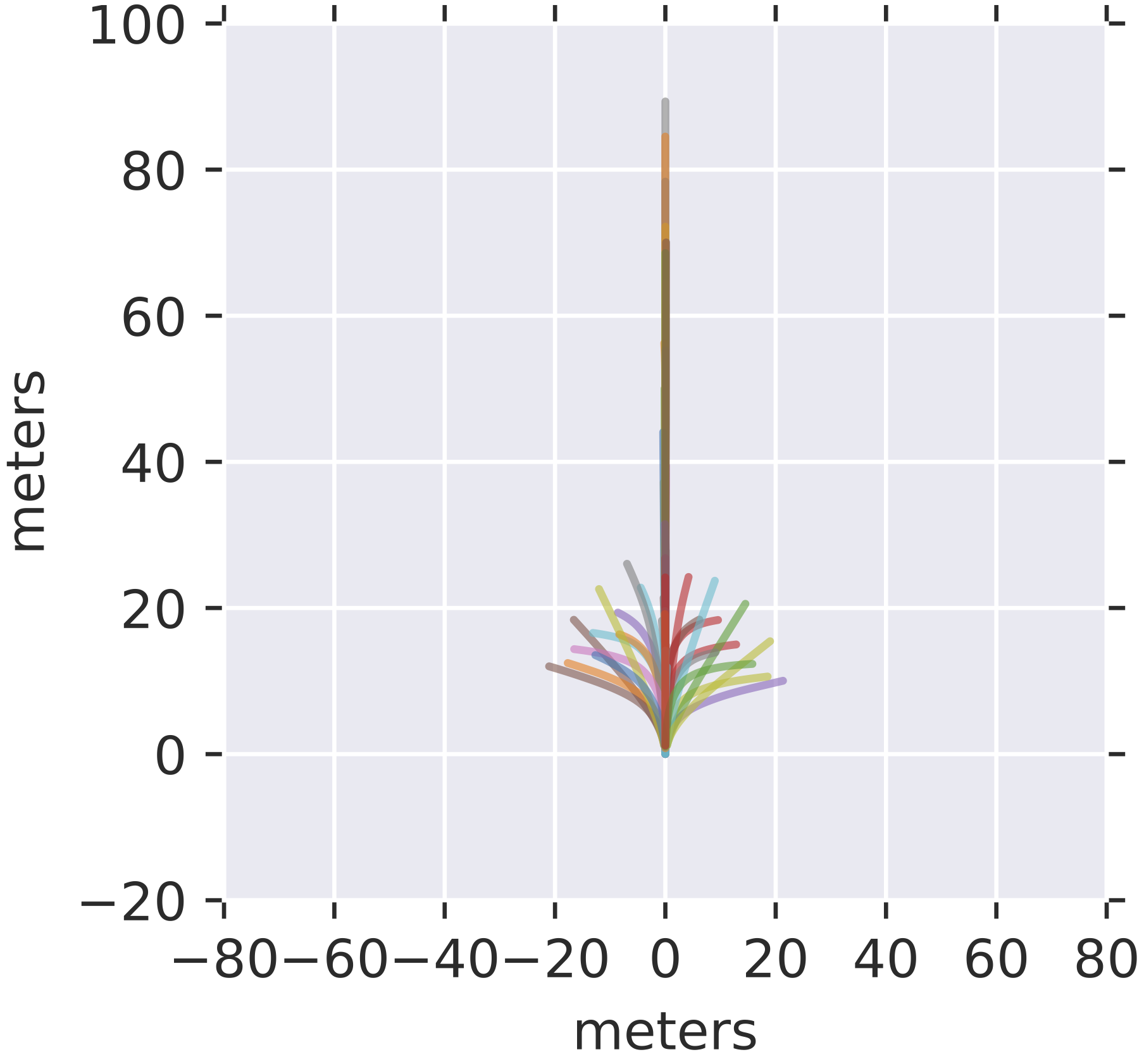}
    \caption{$K=64$}
  \end{subfigure}
  \begin{subfigure}[l]{0.24\linewidth}
    \includegraphics[width=\textwidth]{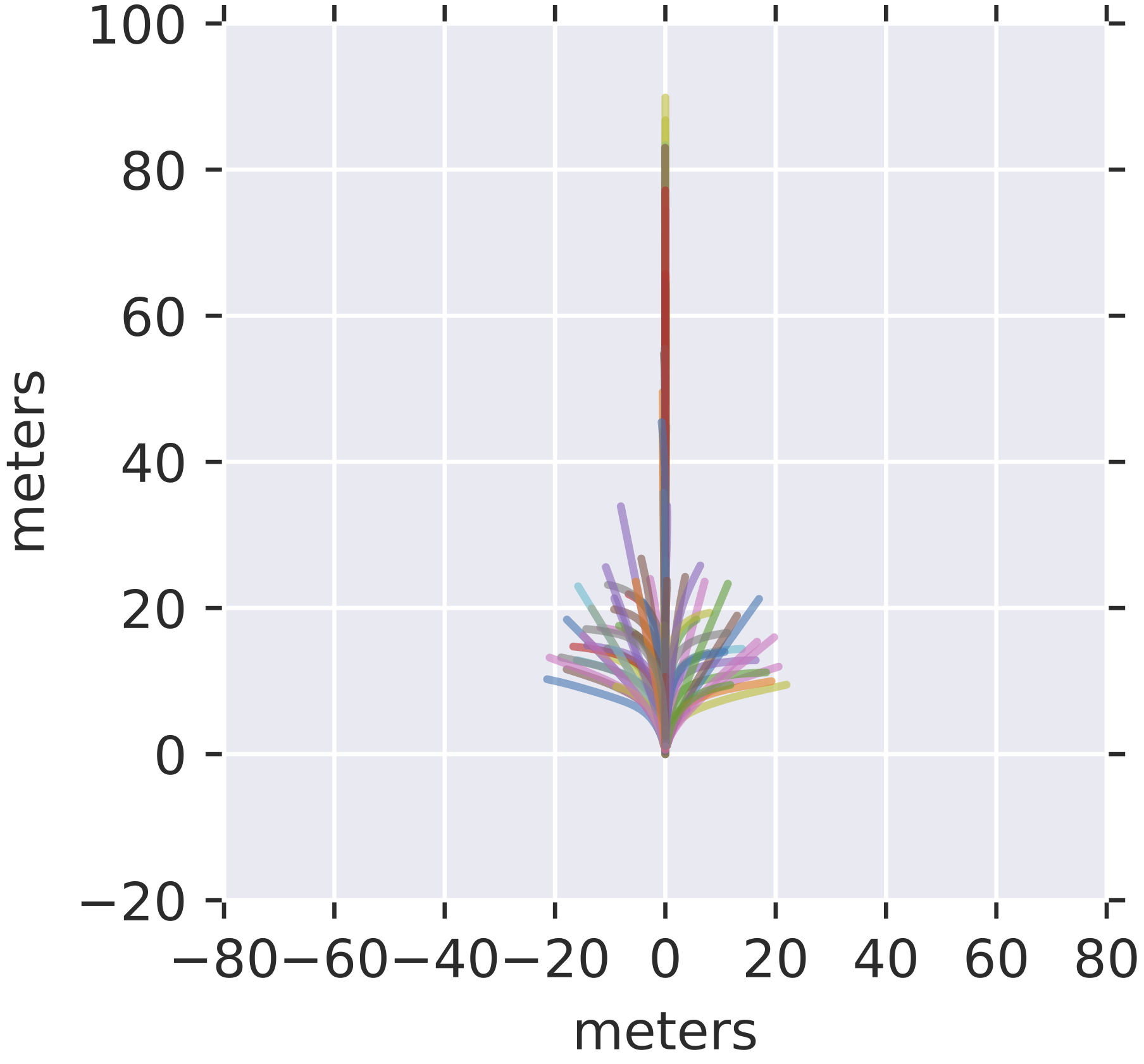}
    \caption{$K=128$}
  \end{subfigure}
  \begin{subfigure}[l]{0.24\linewidth}
    \includegraphics[width=\textwidth]{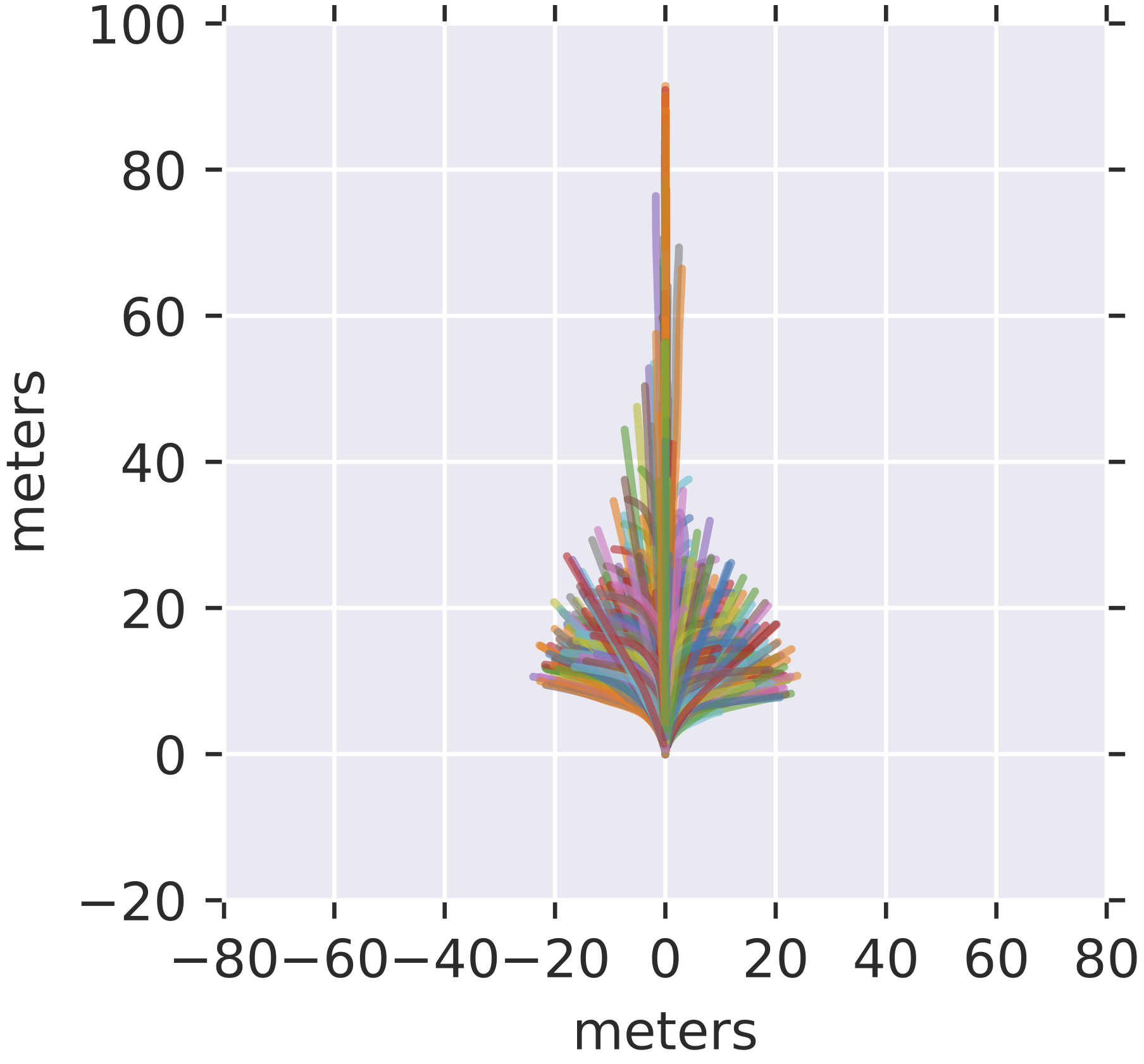}
    \caption{$K=1024$}
  \end{subfigure}
  \caption{Our trajectory library from CARLA's autopilot demonstrations, 4 seconds.}
  \label{fig:trajectory-library}
  \end{figure}

In the absence of scalable global optimizers, we search the trajectory space in Eqn.~(\ref{eq:generic-objective}) by restricting the search space to a trajectory library~\citep{liu2009standing}, $\mathcal{T}_{\mathbf{Y}}$, a finite set of fixed trajectories.
In this work, we perform $K$-means clustering of the expert plan's from the training distribution and keep $64$ of the centroids, as illustrated in Figure~\ref{fig:trajectory-library}.
Therefore we efficiently solve a search problem over a discrete space rather than an optimization problem of continuous variables.
The modified objective is:
\begin{align}
  \by^{\mathcal{G}}_{\text{RIP}}
    \approx \argmax_{\color{matlab-red} \by \in \mathcal{T}_{\mathbf{Y}}} \underset{\btheta \in \text{supp}\big(p(\btheta \vert \mathcal{D})\big)}{\oplus} \log p(\by \vert \mathcal{G}, \bx;\btheta)
\label{eq:RIP-Ty}
\end{align}
Solving for Eqn.~(\ref{eq:RIP-Ty}) results in $\times20$ improvement in runtime compared to the gradient descent alternative.
Although in in-distribution scenes solving Eqn.~(\ref{eq:RIP-Ty}) over Eqn.~(\ref{eq:generic-objective}) does not deteriorate perfomance, in out-of-distribution scenes the trajectory library, $\mathcal{T}_{\mathbf{Y}}$, is not useful.
Therefore in the experiments (c.f. Section~\ref{subsub:online-planning-experiments}) we used online gradient-descent.
Future work lies in developing a hybrid optimization method that takes advantage of the speedup the trajectory library provides without a decrease in performance in out-of-distribution scenarios.

\end{document}